\documentclass[11pt, a4paper]{article}

\usepackage[utf8]{inputenc}
\usepackage{amsmath}        
\usepackage{amssymb}        
\usepackage{amsfonts}       
\usepackage{graphicx}       
\usepackage{float}          
\usepackage[margin=1in]{geometry} 
\usepackage{hyperref}       
\usepackage{url}            
\usepackage{booktabs}       
\usepackage{lmodern}        
\usepackage{tikz}
\usepackage{subcaption}
\usepackage{indentfirst}    
\usepackage{makecell}

\usepackage{siunitx}  
\usepackage{multirow} 
\usepackage{bm}

\usepackage{sansmath}

\usetikzlibrary{
    arrows.meta, 
    decorations.pathreplacing,
    decorations.pathmorphing,
    positioning,
    calc,
    shapes.misc,
    fit,
    shadows,
    patterns,
    shapes.geometric,
    shadows.blur
}

\usepackage[utf8]{inputenc} 
\usepackage[T1]{fontenc}    
\usepackage{amsmath}        
\usepackage{graphicx}       

\usepackage{times}

\newcommand{\Woh}{W_{\mathit{oh}}}
\newcommand{\Wog}{W_{\mathit{og}}}

\newcommand{\genNodeLayer}[7]{
    \foreach \i in {1,...,#4} {
        \pgfmathsetmacro{\xPos}{2 * (\i - 1) + #2} %
        \node[#5] (#1-\i) at ({#2 + \xPos}, {#3}) {$#6_{#7\i}$};
    }
}

\newcommand{\genLeftAlignedNodeLayer}[7]{
    \foreach \i in {1,...,#4} {
        \pgfmathsetmacro{\xPos}{2 * (\i - 1)} %
        \node[#5] (#1-\i) at ({#2 + \xPos}, {#3}) {$#6_{#7\i}$};
    }
}

\newcommand{\genNodeColumnLayer}[7]{
    \foreach \i in {1,...,#4} {
        \pgfmathsetmacro{\yPos}{#3 + 2 * (\i -  1)} 
        \node[#5] (#1-\i) at ({#2}, {\yPos}) {$#6_{#7\i}$};
    }
}

\newcommand{\genFullconnect}[6]{
    \foreach \i in {1,...,#3} {
        \foreach \j in {1,...,#6} {
            \pgfmathsetmacro{\startx}{int(\i + #2 - 1)} %
            \pgfmathsetmacro{\starte}{int(\j + #5 - 1)} %
            \draw[->] (#1-\startx) -- (#4-\starte);
        }
    }
}

\newcommand{\genCausalFullconnect}[6]{
    \foreach \i in {1,...,#3} {
        \foreach \j in {1,...,#6} {
            \pgfmathparse{ifthenelse(\i<=\j, 1, 0)}
            
            \ifnum\pgfmathresult=1
                \pgfmathsetmacro{\startx}{int(\i + #2 - 1)} %
                \pgfmathsetmacro{\starte}{int(\j + #5 - 1)} %
                \draw[->, >=latex] (#1-\startx) -- (#4-\starte);
            \fi
        }
    }
}

\title{\textbf{Rethinking Transformer Connectivity: TLinFormer, A Path to Exact, Full Context-Aware Linear Attention}}
\author{Zhongpan Tang \\ 
        \texttt{tangzhongp@gmail.com}}
\date{\today}

\begin{document}

\maketitle

\begin{abstract}
The Transformer architecture has become a cornerstone of modern artificial intelligence, but its core self-attention mechanism suffers from a complexity bottleneck that scales quadratically with sequence length, severely limiting its application in long-sequence tasks. To address this challenge, existing linear attention methods typically sacrifice model performance by relying on data-agnostic kernel approximations or restrictive context selection. This paper returns to the first principles of connectionism, starting from the topological structure of information flow, to introduce a novel linear attention architecture—\textbf{TLinFormer}. By reconfiguring neuron connection patterns, TLinFormer achieves strict linear complexity while computing exact attention scores and ensuring information flow remains aware of the full historical context. This design aims to bridge the performance gap prevalent between existing efficient attention methods and standard attention. Through a series of experiments, we systematically evaluate the performance of TLinFormer against a standard Transformer baseline on long-sequence inference tasks. The results demonstrate that TLinFormer exhibits overwhelming advantages in key metrics such as \textbf{inference latency}, \textbf{KV cache efficiency}, \textbf{memory footprint}, and \textbf{overall speedup}.
\end{abstract}

\section{Introduction}
\label{sec:introduction}

Since its introduction by Vaswani et al.~\cite{vaswani_attention_2023}, the Transformer architecture has achieved revolutionary breakthroughs in the field of Natural Language Processing (NLP). The core of its success lies in the self-attention mechanism, which dynamically constructs context representations by computing pairwise scores between all elements in a sequence.

However, this powerful expressive capability comes at a high computational cost. The computational and memory complexity of the standard self-attention mechanism is $\mathcal{O}(N^2 d)$, where $N$ is the sequence length and $d$ is the feature dimension. For tasks involving long documents, high-resolution images, or long audio, this quadratic scaling poses a severe performance bottleneck.

To address this challenge, a large number of methods, often referred to as "Efficient Transformers" or "X-formers," have emerged. These methods can be broadly categorized as follows:
\begin{itemize}
    \item \textbf{Sparse Methods}: These methods reduce computation by using pre-defined fixed or dynamic sparse attention patterns. For example, \textbf{Longformer}~\cite{beltagy_longformer_2020} combines local windowed attention with global attention, while \textbf{BigBird}~\cite{zaheer_big_2021} introduces random attention patterns. In recent years, sparse attention has made significant progress in addressing the efficiency problem of long-text modeling. A representative work is NSA (Native Sparse Attention)~\cite{yuan_native_2025}. NSA employs an innovative hierarchical architecture that skillfully combines hardware-aligned block-wise computation with end-to-end trainability, achieving superior performance over full-attention models and several-fold training/inference speedups on multiple benchmarks. However, the design of NSA also brings inherent limitations. The method first uses an MLP to compress the input sequence into a shorter proxy representation, then computes importance scores based on this proxy, and selects the top-n original blocks for subsequent computation. This "compress-then-select" indirect process makes its performance heavily dependent on whether the compression module can effectively preserve key information, posing a risk of an information bottleneck. Furthermore, its performance is also highly dependent on a series of hyperparameters that require fine-tuning, such as compression block size, the number of selected blocks n, and the sliding window size, which undoubtedly increases the complexity of its deployment and optimization across different tasks and model scales.

    \item \textbf{Low-Rank Approximation}: The core idea is to approximate the full attention matrix with a low-rank decomposition. \textbf{Linformer}~\cite{wang_linformer_2020} projects the Key and Value matrices into a lower-dimensional space, achieving linear complexity. However, this low-rank assumption may fail to capture all the complex, high-rank dependencies within a sequence.

    \item \textbf{Kernel Methods}: Represented by \textbf{Performer}~\cite{choromanski_rethinking_2022} and \textbf{Linear Transformer}~\cite{katharopoulos_transformers_2020}, these methods use kernel functions to approximate the Softmax function, thereby avoiding the explicit computation of the $N \times N$ attention matrix. Their significant drawback is that the kernel functions they use are often data-agnostic and fixed, which stands in stark contrast to the dynamic nature of Softmax attention.
\end{itemize}

Although these methods have achieved success in specific scenarios, they universally face a critical trade-off: \textbf{linearization is achieved by introducing architectural or mathematical compromises, which leads to a common performance gap between these efficient models and standard Softmax attention.} These compromises, whether sparse patterns or data-agnostic kernel functions, can be seen as a form of top-down symbolism—that is, optimizing the network by imposing pre-set rules. This deviates from the core philosophy of the attention mechanism: allowing the network to dynamically and unrestrictedly learn relationships between tokens from the data.

\textbf{This paper deviates from the mainstream paradigm of approximating attention. Instead, we pose a more fundamental question: Can we design a novel neural architecture that achieves linear computational complexity without sacrificing the exactness of attention computation or the awareness of the full historical context?} To answer this, we return to the basic principles of connectionism, where intelligence emerges from the bottom-up interactions of simple units. We no longer impose complex mathematical constraints but instead go back to the essence of neural networks—\textbf{the connection structure of neurons}—to solve the long-sequence dilemma.

\textbf{To this end, we propose a new architecture called TLinFormer, which provides a new path to achieving an Exact, Full Context-Aware linear attention.} The core design principle of TLinFormer is \textbf{architectural full information reachability}. Unlike kernel-based methods that sacrifice computational precision through approximations, TLinFormer performs exact Softmax attention calculations within its windows. Concurrently, unlike sparse methods that achieve efficiency by dropping or ignoring parts of the input sequence, TLinFormer's unique connection mechanism ensures that in each generation step, the information flow from the entire historical context is theoretically complete and accessible. It addresses computational efficiency by forcing the model to learn how to compress complete historical information into a fixed-size observation window, rather than taking a "shortcut" by discarding information. We argue that this mandatory information compression, even if it brings a slight performance trade-off, is a key step towards more advanced model intelligence.

TLinFormer elegantly resolves the conflict between global information fusion and computational cost. The core contributions of this paper are as follows:
\begin{enumerate}
    \item \textbf{Strictly Linear Attention}: The computational complexity is strictly linear with respect to the sequence length.
    \item \textbf{Exact Attention Computation}: Unlike kernel-based approximation methods, TLinFormer introduces no mathematical approximations in its attention calculation, always performing the exact Softmax operation. This avoids the performance degradation caused by approximation errors.
    \item \textbf{Full Context Awareness}: Unlike sparse attention methods, TLinFormer's unique connection topology ensures that every generation step can be aware of the information from the entire historical context, architecturally preventing the risk of information loss inherent in token dropping.
    \item \textbf{Architectural Compatibility}: The proposed module can be used as a plug-and-play component to replace standard attention layers in existing Transformer architectures.
    \item \textbf{Superior Inference Efficiency and Cache-Friendliness}: The unique windowed computation pattern leads to excellent cache efficiency, greatly accelerating the autoregressive generation process.
\end{enumerate}

\section{A Connectionist View of the Attention Mechanism}
\label{sec:connection_view}

\subsection{Multilayer Perceptron (MLP)}
A Multilayer Perceptron (MLP) is a fundamental unit for processing feature vectors. In a batch setting, it applies an independent transformation to the features of each sample. Specifically, an MLP module maps an input tensor of shape \texttt{[B, X]} to an output tensor of shape \texttt{[B, O]}, where \texttt{B} is the batch size, and \texttt{X} and \texttt{O} are the input and output feature dimensions, respectively. No information is exchanged between samples within the batch.
\begin{figure}[h!]
  \centering
  \resizebox{0.4\textwidth}{!}{\begin{tikzpicture}[
    neuron/.style={circle, draw, minimum size=1cm},
    input/.style={neuron, fill=green!50},
    hidden/.style={neuron, fill=blue!50},
    output/.style={neuron, fill=red!50},
    >=latex
]

\def\numI{3}
\def\numH{4}
\def\numO{2}

\genNodeColumnLayer{I}{0}{-3}{\numI}{input}{x}{1}
\genNodeColumnLayer{H1}{2}{-4}{\numH}{hidden}{h}{1}
\genNodeColumnLayer{H2}{4}{-4}{\numH}{hidden}{h}{2}
\genNodeColumnLayer{O}{6}{-2}{\numO}{output}{o}{1}

\genFullconnect{I}{1}{\numI}{H1}{1}{\numH}
\genFullconnect{H1}{1}{\numH}{H2}{1}{\numH}
\genFullconnect{H2}{1}{\numH}{O}{1}{\numO}

\end{tikzpicture}}
  \caption{A schematic of a Multilayer Perceptron (MLP), which processes each feature vector independently.}
\end{figure}
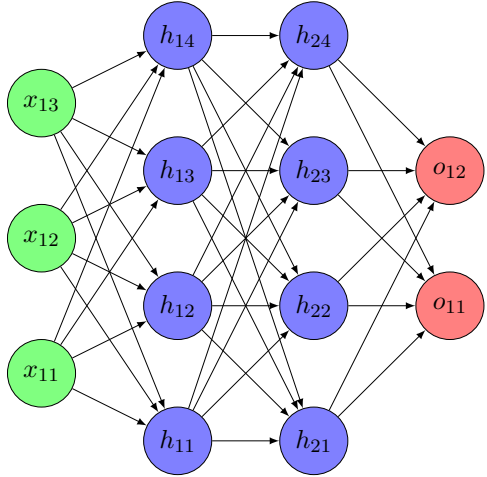

\subsection{Standard Transformer from a Connectionist Perspective}

\begin{figure}[H]
    \centering
    \begin{subfigure}[b]{0.36\textwidth}
        \includegraphics[width=\linewidth]{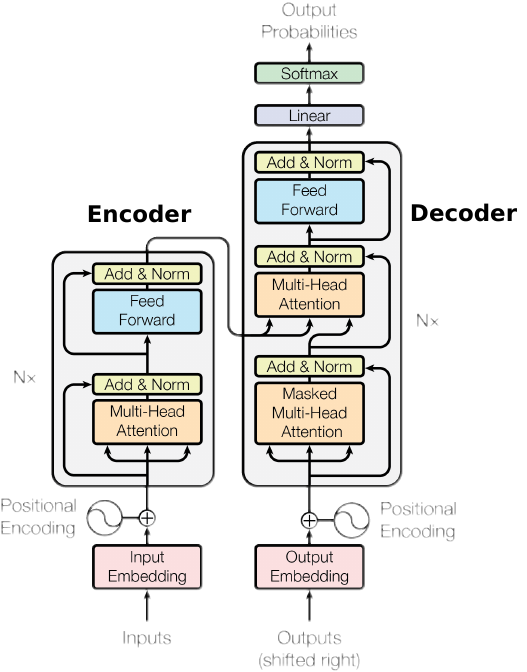}
        \caption{Standard Transformer architecture}
        \label{subfig:standard_transformer}
    \end{subfigure}%
    \hspace{0.1\textwidth}
    \begin{subfigure}[b]{0.36\textwidth}
        \resizebox{\linewidth}{!}{\begin{tikzpicture}[
    neuron/.style={circle, draw, minimum size=1cm},
    input/.style={neuron, fill=green!50},
    hidden/.style={neuron, fill=blue!50},
    output/.style={neuron, fill=red!50},
    memory/.style={neuron, fill=orange!50}, 
    context/.style={neuron, fill=yellow!50}, 
    arrow/.style={->, >=latex, thick}, 
  ]
    
    \def \numX {5}
    \def \numXc {1}
    \def \numXg {2}
    \def \numCH {1}
    \def \numXH {2}
    \def \numO {2}
    
    \genLeftAlignedNodeLayer{X}{0}{0}{\numX}{input}{x}{}
  
    \genLeftAlignedNodeLayer{C}{2}{2}{\numCH}{context}{c}{}
  
    \genLeftAlignedNodeLayer{H1}{6}{4}{\numXH}{hidden}{h}{1}
    \genLeftAlignedNodeLayer{H2}{6}{6}{\numXH}{hidden}{h}{2}
  
    \genLeftAlignedNodeLayer{O}{6}{8}{\numO}{output}{o}{1}
  
    \draw[blue,decorate,decoration={brace,amplitude=20},thick]
        ([yshift=0.1cm]X-1.north) -- ([yshift=0.1cm]X-3.north);

    \genFullconnect{C}{1}{\numXc}{H1}{1}{\numXH}
    \genFullconnect{C}{1}{\numXc}{H2}{1}{\numXH}
    \genFullconnect{C}{1}{\numXc}{O}{1}{\numXH}
    \genCausalFullconnect{X}{4}{\numXg}{H1}{1}{\numXH}
    \genCausalFullconnect{H1}{1}{\numXH}{H2}{1}{\numXH}
    \genCausalFullconnect{H2}{1}{\numXH}{O}{1}{\numXH}
  
  \end{tikzpicture}
  }
        \caption{Conceptual connectivity view}
        \label{subfig:transformer_mlp_view}
    \end{subfigure}%
    \caption{Two views of the standard Transformer architecture. (a) The original modular diagram, from~\cite{vaswani_attention_2023}. (b) A conceptual view from a connectionist perspective for analyzing information flow in autoregressive generation tasks. This view highlights how all historical information is compressed into a single context vector $c_1$ and serves as the starting point for generation.}
    \label{fig:attention_reviews}
\end{figure}
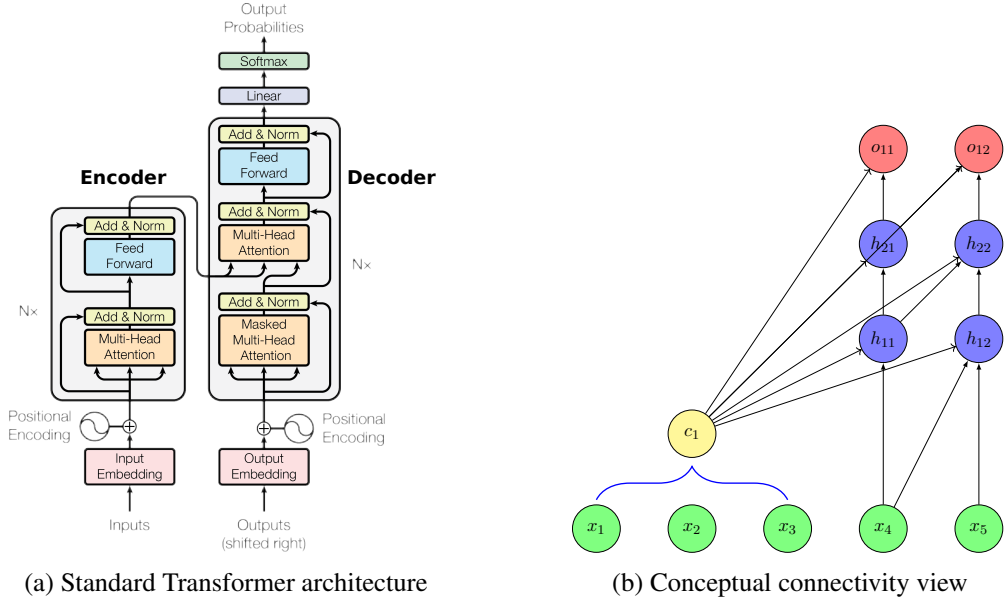

The standard Transformer architecture (Figure~\ref{subfig:standard_transformer}) consists of an encoder and a decoder stack. From a connectivity perspective, the self-attention mechanism can be conceptualized as a generalized linear transformation acting on the sequence axis. With this view, in a generative modeling scenario, we can redraw the standard architecture as shown in Figure~\ref{subfig:transformer_mlp_view}.

In Figure~\ref{subfig:transformer_mlp_view}, historical information ($x_1$ to $x_3$) is compressed by the encoder into a single context representation $c_1$. The generative part ($x_4$ to $x_5$) utilizes this context. From a connectivity perspective, this linear information flow starting from the compressed context $c_1$ presents two significant problems:
\begin{enumerate}
    \item \textbf{Information Bottleneck}: The power of neural networks stems from rich, multi-path information integration. By compressing the entire history into a single vector $c_1$, the standard architecture creates a low-bandwidth bottleneck, severely limiting the depth of interaction between the historical context and the tokens in the generation window. In particular, for certain computational paths, such as $$ x_4 \rightarrow h_{11} \rightarrow h_{21} \rightarrow o_{11} $$, the presence of causal masking forces historical information to flow sequentially along this single path, lacking more direct and richer connections.
    \item \textbf{Structural Fragility}: This reliance on a single context vector makes the network structure fragile. The random deactivation of a single neuron (like $h_{11}$) due to Dropout can sever multiple additional information pathways (e.g., from $c_1$ to $h_{11}$ and then to $h_{22}$). In deep networks, such random interruptions can have a cascading effect, leading to training instability. This fragility may be one of the reasons why early GPT models shifted to the more robust Decoder-only paradigm.
\end{enumerate}

\subsection{The Ideal Connection Structure}
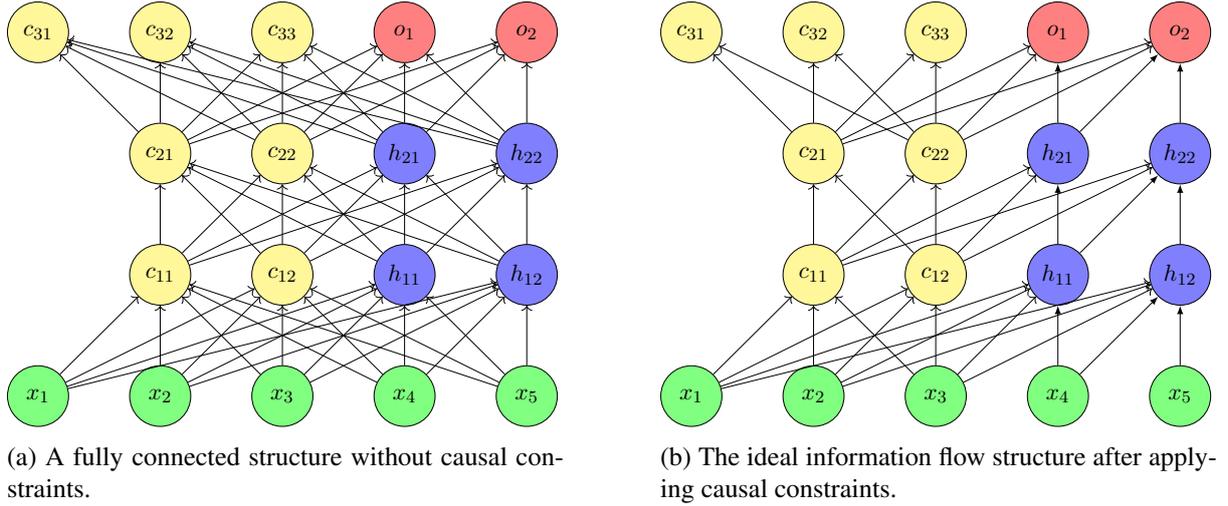
\begin{figure}[H]
  \centering
  \begin{subfigure}[b]{0.46\textwidth}
    \centering
    \resizebox{\linewidth}{!}{\begin{tikzpicture}[
    neuron/.style={circle, draw, minimum size=1cm},
    input/.style={neuron, fill=green!50},
    hidden/.style={neuron, fill=blue!50},
    output/.style={neuron, fill=red!50},
    memory/.style={neuron, fill=orange!50}, 
    context/.style={neuron, fill=yellow!50}, 
    arrow/.style={->, >=latex, thick}, 
]
    
    \def \numX {5}
    \def \numXc {3}
    \def \numXg {2}
    \def \numCH {2}
    \def \numXH {2}
    \def \numO {2}

    \genLeftAlignedNodeLayer{X}{0}{0}{\numX}{input}{x}{}

    \genLeftAlignedNodeLayer{C1}{2}{2}{\numCH}{context}{c}{1}
    \genLeftAlignedNodeLayer{C2}{2}{4}{\numCH}{context}{c}{2}
    \genLeftAlignedNodeLayer{C3}{0}{6}{\numXc}{context}{c}{3}

    \genLeftAlignedNodeLayer{H1}{6}{2}{\numXH}{hidden}{h}{1}
    \genLeftAlignedNodeLayer{H2}{6}{4}{\numXH}{hidden}{h}{2}

    \genLeftAlignedNodeLayer{O}{6}{6}{\numO}{output}{o}{}

    \genFullconnect{X}{1}{\numX}{C1}{1}{\numCH}
    \genFullconnect{X}{1}{\numX}{H1}{1}{\numXH}

    \genFullconnect{C1}{1}{\numCH}{C2}{1}{\numCH}
    \genFullconnect{C1}{1}{\numCH}{H2}{1}{\numXH}   
    \genFullconnect{H1}{1}{\numXH}{C2}{1}{\numCH}
    \genFullconnect{H1}{1}{\numXH}{H2}{1}{\numXH} 

    \genFullconnect{C2}{1}{\numCH}{C3}{1}{\numXc}
    \genFullconnect{C2}{1}{\numCH}{O}{1}{\numO}   
    \genFullconnect{H2}{1}{\numXH}{C3}{1}{\numXc}
    \genFullconnect{H2}{1}{\numXH}{O}{1}{\numO}   

\end{tikzpicture}}
    \caption{A fully connected structure without causal constraints.}
    \label{subfig:full_connection}
  \end{subfigure}
  \hfill
  \begin{subfigure}[b]{0.46\textwidth}
    \centering
    \resizebox{\linewidth}{!}{\begin{tikzpicture}[
    neuron/.style={circle, draw, minimum size=1cm},
    input/.style={neuron, fill=green!50},
    hidden/.style={neuron, fill=blue!50},
    output/.style={neuron, fill=red!50},
    memory/.style={neuron, fill=orange!50}, 
    context/.style={neuron, fill=yellow!50}, 
    arrow/.style={->, >=latex, thick}, 
]
    
    \def \numX {5}
    \def \numXc {3}
    \def \numXg {2}
    \def \numCH {2}
    \def \numXH {2}
    \def \numO {2}

    \genLeftAlignedNodeLayer{X}{0}{0}{\numX}{input}{x}{}

    \genLeftAlignedNodeLayer{C1}{2}{2}{\numCH}{context}{c}{1}
    \genLeftAlignedNodeLayer{C2}{2}{4}{\numCH}{context}{c}{2}
    \genLeftAlignedNodeLayer{C3}{0}{6}{\numXc}{context}{c}{3}

    \genLeftAlignedNodeLayer{H1}{6}{2}{\numXH}{hidden}{h}{1}
    \genLeftAlignedNodeLayer{H2}{6}{4}{\numXH}{hidden}{h}{2}

    \genLeftAlignedNodeLayer{O}{6}{6}{\numO}{output}{o}{}
    
    \genFullconnect{X}{1}{\numXc}{C1}{1}{\numCH}
    \genFullconnect{X}{1}{\numXc}{H1}{1}{\numXH}
    \genCausalFullconnect{X}{4}{\numXg}{H1}{1}{\numXH}

    \genFullconnect{C1}{1}{\numCH}{C2}{1}{\numCH}
    \genFullconnect{C1}{1}{\numCH}{H2}{1}{\numXH}   

    \genCausalFullconnect{H1}{1}{\numXH}{H2}{1}{\numXH} 

    \genFullconnect{C2}{1}{\numCH}{C3}{1}{\numXc}
    \genFullconnect{C2}{1}{\numCH}{O}{1}{\numO}   

    \genCausalFullconnect{H2}{1}{\numXH}{O}{1}{\numO}   

\end{tikzpicture}}
    \caption{The ideal information flow structure after applying causal constraints.}
    \label{subfig:ideal_connection}
  \end{subfigure}
  \caption{Two possible connectivity views. The ideal causal structure (b) ensures maximum information flow between all legitimate nodes.}
\end{figure}

An optimal structure should facilitate comprehensive interaction between historical and generative information. By removing all connections that violate causality from a fully connected graph (Figure~\ref{subfig:full_connection}), we derive the ideal causal information flow structure shown in Figure~\ref{subfig:ideal_connection}.

\section{The TLinFormer Architecture}
\label{sec:architecture}

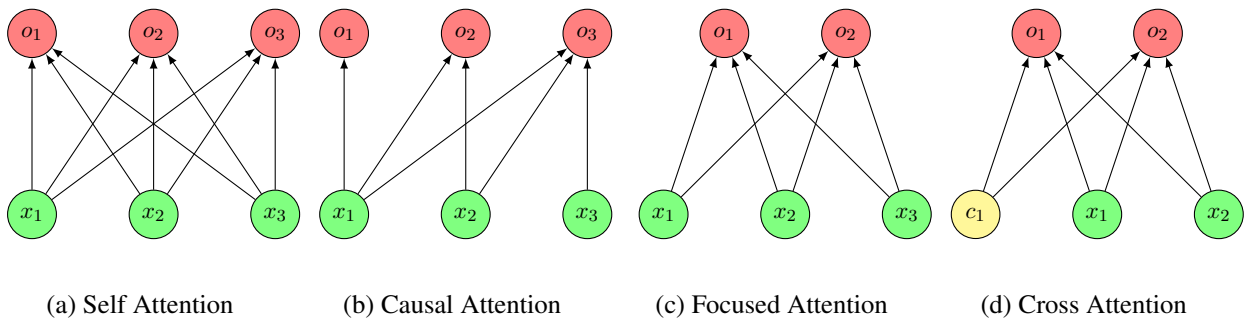
\begin{figure}[H]
    \centering

    \begin{subfigure}[b]{0.22\textwidth}
        \centering
        \begin{tikzpicture}[
    neuron/.style={circle, draw, minimum size=0.8cm}, 
    input/.style={neuron, fill=green!50},
    hidden/.style={neuron, fill=blue!50},
    output/.style={neuron, fill=red!50},
    >=latex,
    scale=0.8, transform shape 
]
    \def\numI{3}
    \def\numO{3}
    \genNodeLayer{I}{0}{0}{\numI}{input}{x}{}
    \genNodeLayer{O}{0}{3}{\numO}{output}{o}{}
    \genFullconnect{I}{1}{\numI}{O}{1}{\numO}
\end{tikzpicture}
        \caption{Self Attention}
        \label{subfig:self_attention}
    \end{subfigure}%
    \hfill
    \begin{subfigure}[b]{0.22\textwidth}
        \centering
        \begin{tikzpicture}[
    neuron/.style={circle, draw, minimum size=0.8cm},
    input/.style={neuron, fill=green!50},
    hidden/.style={neuron, fill=blue!50},
    output/.style={neuron, fill=red!50},
    >=latex,
    scale=0.8, transform shape
]
    \def\numI{3}
    \def\numO{3}
    \genNodeLayer{I}{0}{0}{\numI}{input}{x}{}
    \genNodeLayer{O}{0}{3}{\numO}{output}{o}{}
    \genCausalFullconnect{I}{1}{\numI}{O}{1}{\numO}
\end{tikzpicture}
        \caption{Causal Attention}
        \label{subfig:causal_attention}
    \end{subfigure}
    \hfill
    \begin{subfigure}[b]{0.22\textwidth}
        \centering
        \begin{tikzpicture}[
    neuron/.style={circle, draw, minimum size=0.8cm},
    input/.style={neuron, fill=green!50},
    hidden/.style={neuron, fill=blue!50},
    output/.style={neuron, fill=red!50},
    >=latex,
    scale=0.8, transform shape
]
    \def\numI{3}
    \def\numO{2}
    \genNodeLayer{I}{0}{0}{\numI}{input}{x}{}
    \genNodeLayer{O}{0.5}{3}{\numO}{output}{o}{}
    \genFullconnect{I}{1}{\numI}{O}{1}{\numO}
\end{tikzpicture}
        \caption{Focused Attention}
        \label{subfig:focused_attention}
    \end{subfigure}%
    \hfill
    \begin{subfigure}[b]{0.22\textwidth}
        \centering
        \begin{tikzpicture}[
    neuron/.style={circle, draw, minimum size=0.8cm},
    input/.style={neuron, fill=green!50},
    hidden/.style={neuron, fill=blue!50},
    output/.style={neuron, fill=red!50},
    context/.style={neuron, fill=yellow!50}, 
    >=latex,
    scale=0.8, transform shape
]
    \def\numC{1}
    \def\numI{2}
    \def\numO{2}

    \genNodeLayer{C}{0}{0}{\numC}{context}{c}{}
    \genNodeLayer{I}{1}{0}{\numI}{input}{x}{}
    \genNodeLayer{O}{0.5}{3}{\numO}{output}{o}{}
    \genFullconnect{C}{1}{\numC}{O}{1}{\numO}
    \genFullconnect{I}{1}{\numI}{O}{1}{\numO}
\end{tikzpicture}
        \caption{Cross Attention}
        \label{subfig:cross_attention}
    \end{subfigure}%
    
    \caption{Connection diagrams for the 4 types of Attention mechanisms required in this paper.}
    \label{fig:attention_types}
\end{figure}

To realize the ideal connection structure in Figure~\ref{subfig:ideal_connection}, we need to re-examine the attention mechanism from a new perspective. Its formula is:
\[
\text{Attention}(\bm{Q}, \bm{K}, \bm{V}) = \text{softmax}\left(\frac{\bm{Q} \bm{K^T}}{\sqrt{d_k}}\right) \bm{V}
\]
We no longer view it as a traditional "query-key-value" interaction, but rather as a generalized matrix multiplication or a dynamically weighted fully connected layer acting on the sequence dimension. As shown in Figure~\ref{fig:attention_types}, a standard MLP (i.e., a fully connected layer) uses a fixed weight matrix to transform the features of each token independently, whereas the attention mechanism dynamically generates a data-dependent weight matrix via $QK^T$ and performs a weighted aggregation of information across the sequence dimension. This mechanism is essentially a dynamic information reorganization operation on a tensor of shape [B, L, $d_{\text{model}}$]. Based on this understanding, we can precisely construct our desired information flow by designing and combining different attention patterns.

As shown in Figure~\ref{fig:attention_types}, we can unify several core attention mechanisms under a common framework from a connectivity perspective. The core operation of attention can be abstracted as the interaction between Query ($\bm{Q}$), Key ($\bm{K}$), and Value ($\bm{V}$). Different attention types essentially impose different constraints on the sources of $\bm{Q}, \bm{K}, \bm{V}$ and the connection patterns between them:

\begin{description}
    \item[Self Attention:] 
    As shown in Figure~\ref{subfig:self_attention}, its query, key, and value all come from the same input sequence $\bm{X}$, i.e., $\bm{Q}=\bm{K}=\bm{V}=\bm{X}$. This allows each element in the sequence to attend to all other elements, including itself, thereby building a global contextual representation.

    \item[Causal Attention:]
    As shown in Figure~\ref{subfig:causal_attention}, this is a special form of self-attention mainly used for autoregressive decoding tasks. It modifies the connectivity by applying an upper-triangular mask, ensuring that when computing the output at position $i$, its query $\bm{q}_i$ can only attend to keys $\bm{k}_j$ and values $\bm{v}_j$ at positions $j \le i$. This guarantees that the model does not "see" future information when predicting the future, thus maintaining the causal property of autoregression.

    \item[Focused Attention:]
    As shown in Figure~\ref{subfig:focused_attention}, this is a more generalized form of attention aimed at balancing computational efficiency with information completeness. Its core idea is to assign the responsibility of "information summarization" to a specific subset of the input sequence $\bm{X}$.

    Mathematically, this means the query set $\bm{Q}$ is formed by this subset ($\bm{Q} \subseteq \bm{X}$), while the key set $\bm{K}$ and value set $\bm{V}$ are still formed by the complete input sequence ($\bm{K}=\bm{V}=\bm{X}$). By performing the attention computation $\text{softmax}(\bm{Q}\bm{K}^T)\bm{V}$, this selected query subset $\bm{Q}$ will attend to and aggregate information from the entire sequence $\bm{X}$.
    
    Therefore, although the starting point of the computation (i.e., the number of queries) is sparse, the final output representations (corresponding to the positions of the query subset $\bm{Q}$) have already fused global contextual information. From an information flow perspective, since the complete sequence $\bm{X}$ participates in the information fusion process as $\bm{K}$ and $\bm{V}$, no information is discarded a priori.

    \item[Cross Attention:] 
    As shown in Figure~\ref{subfig:cross_attention}, the cross-attention mechanism handles interactions from \textbf{two different information sources}. In this mode, the query $\bm{Q}$ comes from one sequence (e.g., the input sequence $\bm{X}$), while the key $\bm{K}$ and value $\bm{V}$ come from some intermediate layers, such as $c_{21}~c_{22}$ in Figure~\ref{subfig:ideal_connection}.
\end{description}

A TLinFormer block operates on an input that is partitioned into a historical context window $X_{\mathit{hist}}$ and a generation window $X_{\mathit{gen}}$. The ideal connectivity is constructed layer by layer as follows:
\begin{enumerate}
    \item \textbf{Context Path Encoding}: The historical context $X_{\mathit{hist}}$ is compressed in the first layer through the attention shown in Figure~\ref{subfig:focused_attention}. Subsequent intermediate layers are processed by self-attention layers. In the final layer, the dimensionality is restored through the attention shown in Figure~\ref{subfig:cross_attention} (of course, if stacking multiple TLinFormers is not considered, the computation of the final layer can be omitted).

    \item \textbf{Generation Path Computation}: The generation window $X_{\mathit{gen}}$ is processed in parallel. At each layer $i$, its computation involves two information flows:
    \begin{itemize}
        \item \textbf{Internal Cohesion (Causal Self-Attention)}: A causal self-attention mechanism is applied to the generation window representation from the previous layer ($H_{i-1}$, where $H_0 = X_{\mathit{gen}}$). This allows tokens within the generation window to interact with each other while respecting causal constraints.
        \item \textbf{Context Integration (Cross-Attention)}: Historical information is fused into the generation window using a cross-attention mechanism. The queries come from the generation path ($H_{i-1}$), while the keys and values come from the corresponding layer of the context path ($C_{i-1}$) or $X_{\mathit{hist}}$.
        \item The results of these two attention mechanisms are combined and passed through a feed-forward network (FFN) to produce the output of the current layer, $H_i$.
    \end{itemize}
\end{enumerate}

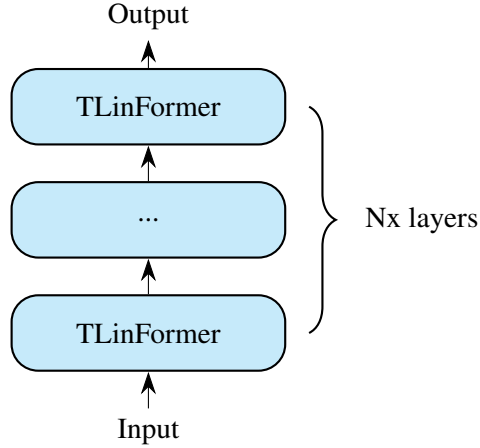
\begin{figure}[H]
    \centering
    \begin{tikzpicture}
    \pgfmathsetmacro {\height} {1}
    \pgfmathsetmacro {\gap} {0.5}
    \pgfmathsetmacro {\num} {3}
    \pgfmathsetmacro {\yposFinal} {\height * \num + (\num + 1) * \gap}

    \foreach \i in {1,...,\num} {
        \pgfmathsetmacro{\ypos}{(\height + \gap) * (\i - 1) + \height/2};

        \pgfmathparse{ifthenelse(\i==2, 1, 0)}
        \ifnum\pgfmathresult=1
            \def\mtext{...}       
        \else
            \def\mtext{TLinFormer}  
        \fi
        \node[draw, rounded corners=10pt, fill=cyan!20, thick, minimum width=3.6cm, minimum height=\height cm] (T\i) at (2,\ypos cm) {\mtext};
    }
 
    \foreach \i in {1,...,2} {
        \pgfmathtruncatemacro{\nexti}{\i + 1}   
        \draw[-{Stealth[length=3mm, width=2mm]}] (T\i.north)  -- (T\nexti.south) ;
    }
    
    \node[below] (TEXT-1) at (2,-\gap cm) {Input};
    \node[below] (TEXT-2) at (2,\yposFinal cm) {Output};

    \draw[-{Stealth[length=3mm, width=2mm]}] (TEXT-1.north)  -- (T1.south) ;
    \draw[-{Stealth[length=3mm, width=2mm]}] (T\num.north)  -- (TEXT-2.south) ;
    
    \draw[black,decorate,decoration={brace,amplitude=10},thick]
    ([xshift=0.3cm]T\num.east)  -- ([xshift=0.3cm]T1.east) node[midway, right=0.6cm] {Nx layers};

\end{tikzpicture}
    \caption{Schematic of a stacked TLinFormer network architecture.}
    \label{fig:stacked_tlin}
\end{figure}

A TLinFormer block can either function as a standalone module or be stacked repeatedly to form a deep network. When used alone, the attention in the final layer of the historical window in Figure~\ref{subfig:ideal_connection} ($C_3$) can be omitted. When multiple blocks are stacked, as shown in Figure~\ref{fig:stacked_tlin}, the output of each layer serves as the input to the next, thereby constructing a standard deep Transformer decoder structure.

\section{Linear Complexity Analysis}
\label{sec:complexity}

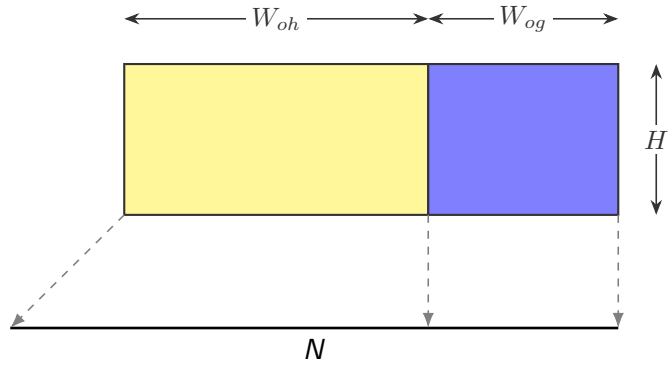
\begin{figure}[H]
    \centering
    \begin{tikzpicture}[
    font=\sansmath\sffamily, 
    rect_style/.style={draw=black!80, line width=0.8pt},             
    proj_line/.style={-Latex, line width=0.6pt, gray, dashed},      
    dim_arrow/.style={<->, >=Stealth, line width=0.6pt, black!80},   
    dim_label/.style={fill=white, inner sep=2pt, font=\small}        
]
    \colorlet{colorA}{yellow!50}
    \colorlet{colorB}{blue!50}

    \def\woh{4}
    \def\w{2.5}
    \def\h{2}
    \def\baselineY{-1.5}
    \def\dimlineY{2.6}

    \begin{scope}[blur shadow={shadow opacity=25, shadow blur steps=5}]
        \filldraw[fill=colorA, rect_style] (0,0) rectangle (\woh, \h);
        \filldraw[fill=colorB, rect_style] (\woh,0) rectangle (\woh + \w, \h);
    \end{scope}

    \draw[line width=1pt] (-1.5, \baselineY) -- (\woh + \w, \baselineY) 
        node[midway, below]{$N$};

    \draw[proj_line] (0,0) -- (-1.5, \baselineY);
    \draw[proj_line] (\woh,0) -- (\woh, \baselineY);
    \draw[proj_line] (\woh + \w, 0) -- (\woh + \w, \baselineY);

    \draw[dim_arrow] (0, \dimlineY) -- (\woh, \dimlineY)
        node[midway, dim_label] {$W_{oh}$};
    \draw[dim_arrow] (\woh, \dimlineY) -- (\woh + \w, \dimlineY)
        node[midway, dim_label] {$W_{og}$};

    \pgfmathsetmacro {\startXh} {\woh + \w + 0.5}
    \draw[dim_arrow] (\startXh, 0) -- (\startXh, \h)
        node[midway, dim_label] {$H$};

\end{tikzpicture}
    \caption{Schematic of windowed computation.}
    \label{fig:windowed_div}
\end{figure}

The input sequence is divided into a historical context window $\Woh$ and a generation window $\Wog$. Let the total sequence length be $N$, the feature dimension be $D$, and the number of intermediate self-attention layers within a TLinFormer block be $H$.

The detailed derivation discussed below can be found in Appendix~\ref{app:complexity_derivation}.

\subsection{Cache Miss}

A \texttt{Cache Miss} refers to an event where pre-computed results cannot be reused, necessitating a full computation from scratch. It primarily occurs under the following two conditions:
\begin{enumerate}
    \item \textbf{During the training phase}: Since the historical context differs for each training batch, the caching mechanism is not applicable. Thus, every forward pass can be considered a \texttt{Cache Miss}.
    \item \textbf{During the inference phase}:
    \begin{itemize}
        \item When generating the initial token for a given context.
        \item At the moment of generating the first new token immediately after the historical context window has been updated (i.e., slided).
    \end{itemize}
\end{enumerate}

The computational cost of a cache miss is equivalent to that of a full forward pass with caching disabled.

The total computational cost is the sum of the costs for the context window and the generation window. The key insight is that the total cost is an exact linear function of the total sequence length, of the form:
\begin{equation}
\text{Total Cost} = C_1 \cdot N + C_0
\label{eq:total_cost_simple_cache_miss}
\end{equation}
where the slope and intercept are constants determined by the model hyperparameters:
\begin{align}
C_1 &= D \cdot (2\Woh + \Wog) \\
C_0 &= D \left[ H(\Woh^2 + \Wog^2 + \Wog\Woh) + \Wog^2 - \Wog\Woh \right] \label{eq:intercept_c0_updated}
\end{align}
This linear relationship arises from the algebraic simplification of the full cost expression, whose complete form is (see Appendix~\ref{app:complexity_derivation} for detailed derivation):
\begin{equation}
\text{Total Cost} = D \left[ N(2\Woh + \Wog) + H(\Woh^2 + \Wog^2 + \Wog\Woh) + \Wog^2 - \Wog\Woh \right]
\label{eq:total_cost_full_cache_miss}
\end{equation}

Since $D, \Woh, \Wog$, and $H$ are all fixed after training, \textbf{the computational complexity of TLinFormer is strictly linear with respect to the sequence length $N$}.

\subsection{Cache Hit}

A \texttt{Cache Hit} occurs exclusively during autoregressive inference. It refers to the event of generating any subsequent token, after the first one, within a single generation cycle (i.e., while the historical context window remains static).

The total computational cost is given by:
\begin{equation}
\text{Total Cost} = D N - D \Wog + (H + 1) D \Woh + (H+2)D \Wog^2
\label{eq:total_cost_full_cache_hit}
\end{equation}

\subsection{Complexity Summary}

The computational complexity of TLinFormer exhibits a dual-mode characteristic tightly coupled with its cache state, which forms the core of its high-efficiency inference capabilities.

\begin{itemize}
    \item \textbf{On a Cache Miss}, for instance during training or when generating an initial token, the model's computational cost is \textbf{strictly linear} with respect to the total sequence length $N$, with a complexity of $\mathcal{O}(N)$. This establishes the upper bound for the model's single-step computational overhead.

    \item \textbf{On a Cache Hit}, i.e., when generating subsequent tokens during autoregressive inference, the model's computational cost remains \textbf{strictly linear} with respect to the total sequence length $N$, but with a significantly reduced slope.
\end{itemize}

This allows TLinFormer to maintain the overhead of most generation steps at a relatively low level when processing long sequences, thereby achieving an order-of-magnitude speedup in inference.

\section{Model Properties and Discussion}
\label{sec:discussion}

\subsection{Training Process and Window Allocation}
\label{sec:discussion_of_windows}
Foundational principles from information theory and compressed sensing, notably the empirical guideline \( n > C \log N \), establish that high-dimensional signals often reside on a low-dimensional manifold~\cite{candes_robust_2004, abo-zahhad_compressive_2015}.
Here, \(n\) is the dimension of the compressed representation required to faithfully reconstruct a signal of original dimension \(N\).
This theoretical underpinning suggests that for processing long sequences, a remarkably small context window can be sufficient.
For instance, to capture the essential information of a sequence with \(N = 10^7\) tokens, a compressed representation of dimension \( n \approx 134 \) (for \( C \approx 8.33 \)) could theoretically suffice.
For analytical tractability and to establish a robust baseline, we adopt a fixed-size historical context window (e.g., 256 tokens) in this work.
The benchmarks established with this approach can be viewed as a strong baseline for future investigations into dynamic windowing schemes.

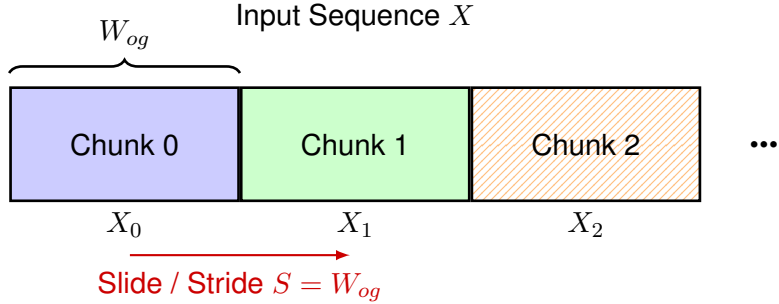
\begin{figure}[H]
    \centering
    \begin{tikzpicture}[
        font=\sffamily,
        >=latex,
        block/.style={draw, rectangle, minimum height=1.5cm, minimum width=3cm, line width=1pt},
        brace/.style={decorate, decoration={brace, amplitude=5pt, raise=5pt}, line width=1pt}
    ]

    \node[block, fill=blue!20, label=below:\(X_0\)] (block0) {Chunk 0};

    \node[block, fill=green!20, label=below:\(X_1\), right=0cm of block0] (block1) {Chunk 1};

    \node[block, fill=orange!20, pattern=north east lines, pattern color=orange!50, label=below:\(X_2\), right=0cm of block1] (block2) {Chunk 2};

    \node[right=0.5cm of block2, font=\Large] (ellipsis) {\textbf{...}};
    
    \node[above=0.6cm of block1] {Input Sequence \(X\)};

    \draw[brace] (block0.north west) -- (block0.north east) node[midway, above=10pt] {$W_{og}$};
    
    \draw[->, thick, red!80!black, shorten <=2pt, shorten >=2pt] 
        ([yshift=-20pt]block0.south) -- 
        node[midway, below=2pt, text=red!80!black] {Slide / Stride \(S = W_{og}\)} 
        ([yshift=-20pt]block1.south);

    \end{tikzpicture}
    
    \caption{Sliding window information processing flow during training.}
    \label{fig:wog_slide}
\end{figure}

The model processes long sequences in chunks. During training, it employs a sliding window mechanism. First, it processes the interval $[0, \Wog]$, where the historical context is empty. Then, the window slides by a distance of $\Wog$, and the model processes the interval $[\Wog, 2\Wog]$, using the first chunk $[0, \Wog]$ as its historical context. The window slides again by $\Wog$, and the model processes the interval $[2\Wog, 3\Wog]$, using the first two chunks $[0, 2\Wog]$ as its historical context. This process continues until the entire sequence is processed. The outputs of each generation window are concatenated to form the final output sequence for loss computation.

This design introduces a critical trade-off between the historical window $\Woh$ and the generation window $\Wog$, which is most evident in the following two extreme cases:
\begin{itemize}
    \item \textbf{An excessively small historical window}: The model will encounter a severe information bottleneck when handling generation tasks. Particularly at the beginning of the generation window, the path for information to travel from the bottom layers to the top becomes extremely narrow, severely weakening the model's ability to capture and utilize long-range dependencies, which contradicts our design philosophy.
    \item \textbf{An excessively large historical window}: This enhances the model's ability to process and understand long-term historical context, which usually leads to improved accuracy. However, under the constraints of limited hardware resources (such as VRAM), an overly large historical window will inevitably squeeze the space available for the generation window. In an extreme case, if the generation window is forced to shrink to a very small size (e.g., 1), computational efficiency will drop sharply: this not only drastically slows down training but also requires frequent reconstruction of the entire historical context cache during inference, thereby significantly increasing latency.
\end{itemize}

In summary, the optimal ratio between the historical and generation windows is not fixed but is closely tied to the specific task and hardware resources. This provides users with a flexible and necessary trade-off space between the depth of the model's contextual understanding and the computational efficiency of training and inference.

\subsection{Superior Cache Efficiency During Inference}
\label{sec:cache_efficiency}

\subsubsection{Lower Memory Footprint}
The KV cache of a standard Transformer scales linearly with the entire sequence length $L$, as shown in Equation~\eqref{eq:kv_cache_memory}, which often becomes a memory bottleneck.
\begin{equation}
M_{\text{transformer}} = 2  \cdot B \cdot L \cdot d_{\text{model}} \cdot P_{\text{bytes}} \cdot N_{\text{layers}}
\label{eq:kv_cache_memory}
\end{equation}

Here, $N_{\text{layers}}$ is the number of layers, $B$ is the batch size, $d_{\text{model}}$ is the hidden dimension, and $P_{\text{bytes}}$ is the byte precision.

TLinFormer offers significant advantages in autoregressive inference. The key-value (KV) cache associated with the historical context window ($\Woh$) remains static as long as the context itself does not change. A cache invalidation and re-computation event only occurs after $\Wog$ new tokens have been generated and the context window needs to be updated.

To quantify this advantage, we perform an approximate analysis. Assume a standard Transformer with $N_{\text{layers}}$ layers has a comparable total computational depth to a TLinFormer with an internal depth of $H$ that is stacked multiple times. In long-sequence scenarios ($N \gg \Woh, \Wog$), the main cache overhead of TLinFormer comes from the few layers that interact with the entire historical sequence. Ultimately, $M_{\text{TLinFormer}}$ can be approximated as:
\begin{equation}
M_{\text{TLinFormer}}  \approx \frac{2 \cdot B \cdot L \cdot d_{\text{model}} \cdot P_{\text{bytes}} \cdot N_{\text{layers}}}{H + 2}
\label{eq:tlin_cache_approx}
\end{equation}

Therefore, the ratio of memory footprint between TLinFormer and a standard Transformer can be expressed as:
\begin{equation}
\frac{M_{\text{TLinFormer}}}{M_{\text{transformer}}} \approx = \frac{1}{H+2}
\label{eq:memory_ratio}
\end{equation}

For a model with a total depth of 100 layers, for instance, if TLinFormer is configured with an internal depth of $H=8$, its cache footprint is theoretically reduced to approximately $\frac{1}{10}$ of a standard model's. If stacking of TLinFormer blocks is forgone and it is directly configured with $H=98$, the cache footprint can theoretically be reduced to approximately $\frac{1}{100}$ of the standard model's. This order-of-magnitude memory optimization significantly lowers the hardware barrier for processing ultra-long sequences and represents another key practical advantage of this architecture.

\subsubsection{Better Time Acceleration: From Global Re-computation to Localized Updates}

A standard autoregressive Transformer (Decoder-only) faces a fundamental efficiency bottleneck during inference. To clarify the advantages of TLinFormer, we first analyze the caching dilemma of the standard model:

\begin{itemize}
    \item \textbf{The Caching Dilemma of Standard Transformers:} After a new token is generated at each step, it is appended to the end of the input sequence. Due to its \textbf{Global Attention} mechanism, every token in the sequence needs to interact with all historical tokens, including the new one. This leads to two severe overheads:
    \begin{enumerate}
        \item \textbf{Computational Redundancy:} Although past keys (K) and values (V) can be cached, the attention weight matrices in all layers must be \textbf{completely re-computed} based on the grown, full sequence.
        \item \textbf{Memory Bandwidth Bottleneck:} More critically, to update the cache, a memory-intensive concatenation operation (like \texttt{torch.cat}) must be performed at every decoding step and in every layer, merging the new token's KV vectors with the existing KV cache. As the sequence grows longer, the overhead of this operation can offset or even exceed the time saved from re-computation, causing the caching mechanism to become \textbf{almost ineffective} at a high level (Figure~\ref{fig:inference_analysis}(c)).
    \end{enumerate}
    
    \item \textbf{TLinFormer's Efficient Caching Strategy:} TLinFormer's windowed architecture fundamentally changes this dynamic, achieving a \textbf{high degree of localization} in computation:
    \begin{enumerate}
        \item \textbf{Static Caching of Historical Context:} During the multiple generation steps where the context window does not slide, all computations related to the historical context window ($\Woh$), including its internal KV cache and the intermediate representations processed by each layer, are \textbf{completely static}. These results can be computed once and reused, completely avoiding repetitive processing of the long historical sequence.
        \item \textbf{Constrained Computation in the Generation Window:} The computational focus is strictly confined to the generation window ($\Wog$).
            \begin{itemize}
                \item When the generation window interacts with the historical window, since the representations of the historical window are static, most of the computation (except for the last new token) can also be reused.
                \item Although the causal self-attention within the generation window needs to be updated, its computational cost is strictly limited to the range of $\mathcal{O}(\Wog^2)$.
            \end{itemize}
    \end{enumerate}
    \label{enum:discussion_of_cache}
\end{itemize}

Therefore, when the total sequence length $N$ is much larger than the window size $\Wog$, the marginal computational cost of generating each step is primarily determined by the interaction of the new token with the historical sequence of length $N$, with a complexity of approximately $\mathcal{O}(N)$. This results in the total inference time exhibiting a \textbf{linear growth with a very low slope} as the sequence length increases. This theoretical analysis is in perfect agreement with the excellent and stable performance of TLinFormer observed in Figure~\ref{fig:inference_analysis}(b) when the cache is hit, validating the extreme efficiency of its caching mechanism.

\section{Experiments}
\label{sec:experiments}

In this section, we will validate the effectiveness of TLinFormer through a series of experiments. We first detail the experimental setup, including the comparison baselines, model configurations, and evaluation metrics, to ensure the fairness and reproducibility of the experiments. Subsequently, we will present and conduct an in-depth analysis of the main results on the wikitext-103-v1 benchmark.

\subsection{A Note on Long-Context Retrieval Tasks}
The "Needle in a Haystack" benchmark is widely used to evaluate the long-context retrieval capabilities of Large Language Models (LLMs). However, this test primarily measures the complex instruction-following and long-range dependency capabilities that emerge after pre-training on large-scale, diverse corpora. The core contribution of the TLinFormer architecture proposed in this paper is to validate a fundamental improvement in computational and memory efficiency. Due to the limitations of our model scale (41M parameters) and training data (wikitext-103-v1), its design goal is not to replicate the full range of emergent abilities of large-scale LLMs. Therefore, we consider the "Needle in a Haystack" test to be orthogonal to the core objective of this paper, which is to validate architectural efficiency, and thus have not included it in our main evaluation scope. Extending the TLinFormer architecture to larger-scale models to explore its potential in complex retrieval tasks is a promising direction for future research.

\subsection{Implementation Details}
\label{ssec:implementation_details}

\paragraph{Hardware Environment:}
To ensure consistency, all our training, inference, and testing were performed on a single, consumer-grade hardware platform. This platform is configured as follows: one \textbf{NVIDIA GeForce RTX 4090 GPU} (24 GB VRAM), an \textbf{AMD EPYC 7543 CPU}, and 62 GB of system memory.

\paragraph{Software Stack:}
The software stack was consistent across both environments: \textbf{Python 3.12.11}, \textbf{PyTorch 2.7.1}, \textbf{CUDA 12.6}, \textbf{cuDNN 9.5.1}, \textbf{Hugging Face Transformers (v4.55.2)}, and the operating system was \textbf{Ubuntu 22.04.5 LTS}.

\subsubsection{Principle of Fair Comparison and Model Configuration}
\label{sssec:fair_comparison_and_config}

To ensure a fair and meaningful comparison between TLinFormer and the standard Transformer baseline, all experiments adhere to the principle of \textbf{parameter count parity}. The core innovation of TLinFormer lies in the \textbf{reorganization} of information flow, rather than the introduction of new parameterized components. It is essentially a \textbf{topological reconstruction} of standard Transformer modules. Therefore, as long as the total computational depth of a stacked TLinFormer model matches the number of layers in a standard Transformer model, their \textbf{total parameter counts are identical}. This allows us to attribute any performance differences solely to the superiority of the architectural design.

In this experiment, we use a small-scale model configuration with approximately \textbf{41M} parameters. Both the baseline model and our TLinFormer use the same core hyperparameters:
\begin{itemize}
    \item \texttt{vocab\_size}: 50257 (consistent with GPT-2)
    \item \texttt{n\_embd} (embedding dimension): 432
    \item \texttt{n\_head} (number of attention heads): 12
    \item \texttt{n\_transformer\_block} (equivalent total depth): 8
\end{itemize}
For the baseline model, this is a standard 8-layer decoder-only Transformer. For TLinFormer, this equivalent depth of 8 transformations is achieved by stacking 2 TLinFormer blocks, with each block having an internal depth hyperparameter $H=2$.

The training parameters, such as learning rate, are identical for all models. The equivalent batch size is set to 256 (achieved through gradient accumulation).

\subsubsection{Dataset and Evaluation Metrics}
\label{sssec:dataset_and_metrics}

We use the wikitext-103-v1 dataset from the Hugging Face repository `Salesforce/wikitext` for all experiments. This dataset contains approximately 120 million tokens. The model's performance is evaluated by its perplexity (PPL) on the validation set, where lower values indicate better performance.

\subsubsection{Baselines and Model Variants}
\label{sssec:baselines}

We compare TLinFormer with a standard decoder-only Transformer baseline model. To evaluate the performance of the models under different configurations, we trained multiple variants for both architectures. The naming convention for each variant is explained below:

\begin{description}
    \item[\texttt{Base XXX:}] 
    Represents the standard Transformer baseline model. The suffix \texttt{XXX} indicates the sequence length used during its training. For example, \texttt{Base 1K} refers to the baseline model trained with a sequence length of 1K.

    \item[\texttt{TLinFormer XXX-YYY-ZZZ:}]
    Represents our TLinFormer model, with its name composed of three parameters:
    \begin{itemize}
        \item \texttt{XXX}: The total sequence length used during training.
        \item \texttt{YYY}: The total length of the core observation window, i.e., $W_{total} = \Woh + \Wog$.
        \item \texttt{ZZZ}: The ratio of the historical context observation window to the total observation window, i.e., $\Woh/W_{total}$.
    \end{itemize}
    For example, \texttt{TLinFormer 2K-512-0.5} represents a TLinFormer model trained with a sequence length of 2K, a total observation window length of 512, and a historical context window length set to half of the total window ($0.5 \times 512 = 256$).
\end{description}

\subsection{Training Results and Analysis}
\label{ssec:results}

\begin{figure}[H]
    \centering
    \includegraphics[width=0.8\textwidth]{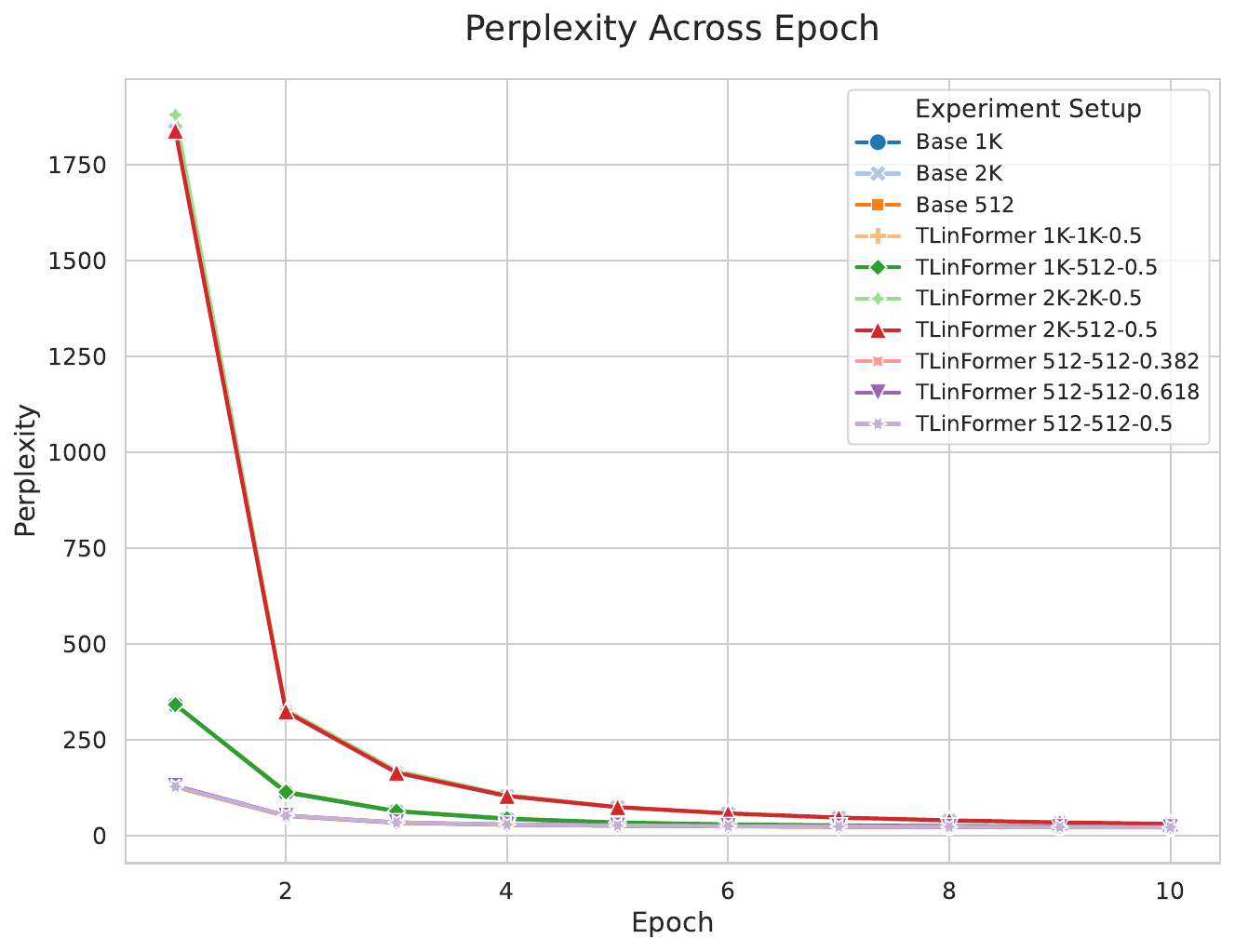} 
    \caption{Perplexity (PPL) of each model over training epochs.}
    \label{fig:ppl_of_tests}
\end{figure}

Table~\ref{tab:ppl_across_epochs} and Figure~\ref{fig:ppl_of_tests} show the perplexity of all model variants on the wikitext-103-v1 validation set as training progresses. We can observe several key points from the results:

\begin{enumerate}
    \item \textbf{Comparable Performance under Equivalent Configuration:}
    First, we verify that the architectural reconstruction of TLinFormer does not introduce performance degradation under a baseline configuration. As shown in Table~\ref{tab:ppl_across_epochs}, when TLinFormer's observation window length is identical to the baseline model's context length (e.g., comparing \texttt{Base 1K} with \texttt{TLinFormer 1K-1K-0.5}), their final perplexities (PPL) are almost identical (22.5 vs 22.7). This confirms that TLinFormer's core design successfully reorganizes the information flow without sacrificing the model's fundamental performance.

    \item \textbf{Controllable Performance Trade-off from Information Compression:}
    Next, we investigate the performance impact of TLinFormer's core "forced compression" mechanism. The experimental results show that when TLinFormer uses an observation window shorter than the sequence length (e.g., \texttt{Base 1K} vs. \texttt{TLinFormer 1K-512-0.5}), its performance exhibits a slight and expected gap compared to the baseline. This performance trade-off is a direct consequence of the model being forced to compress and abstract longer historical information within a limited observation window. We infer that this gap can be further narrowed with extended training. More importantly, we believe this architectural information bottleneck is key to eliciting more advanced intelligent behaviors from the model, a point we will explore in Section~\ref{sec:compression_as_intelligence}.

    \item \textbf{High Robustness to Core Hyperparameters:}
    Finally, we conducted an ablation study on TLinFormer's key hyperparameters (e.g., $\Woh/W_{\text{total}}$). Under the \texttt{TLinFormer 512-512-X} configurations, as shown in Table~\ref{tab:ppl_across_epochs}, despite significant variations in the hyperparameter values, the final PPL of all variants remained stable within a very narrow range around 21.9. This strongly demonstrates that TLinFormer's performance advantage stems from its robust core architectural design, rather than fine-tuning of specific hyperparameters, which greatly enhances its reliability and ease of use in practical applications.
\end{enumerate}

\begin{table}[H]
  \centering
  \caption{Perplexity (PPL) on the wikitext-103-v1 validation set. Lower is better.}
  \label{tab:ppl_across_epochs}
  \sisetup{table-format=4.1}
  \resizebox{\textwidth}{!}{
  \begin{tabular}{@{}l S S S S S S S S S S@{}}
      \toprule
      \multirow{2}{*}{\textbf{Experiment}} & \multicolumn{10}{c}{\textbf{Epoch}} \\
      \cmidrule(l){2-11}
      & {1} & {2} & {3} & {4} & {5} & {6} & {7} & {8} & {9} & {10} \\
      \midrule
      Base 1K                      & 341.5  & 112.5 & 63.4  & 43.5  & 33.0 & 28.3 & 25.8 & 24.2 & 23.3 & 22.5 \\
      TLinFormer   1K-512-0.5     & 341.8  & 114.0 & 64.0  & 44.5  & 34.3 & 29.4 & 26.8 & 25.0 & 23.8 & 23.0 \\
      TLinFormer   1K-1K-0.5       & 340.7  & 115.0 & 64.3  & 44.6  & 34.2 & 29.0 & 26.3 & 24.6 & 23.5 & 22.7 \\
      \midrule
      Base 2K                      & 1839.4 & 321.8 & 162.9 & 102.9 & 73.9 & 57.5 & 46.6 & 38.8 & 32.7 & 29.5 \\
      TLinFormer   2K-512-0.5     & 1839.7 & 324.0 & 164.6 & 103.5 & 74.3 & 58.2 & 47.2 & 40.1 & 34.5 & 30.9 \\
      TLinFormer   2K-2K-0.5       & 1881.0 & 328.5 & 169.4 & 106.0 & 74.6 & 57.7 & 46.8 & 38.9 & 33.7 & 29.8 \\
      \midrule
      Base 512                    & 126.8  & 51.0  & 33.4  & 28.1  & 25.7 & 24.2 & 23.3 & 22.6 & 22.0 & 21.6 \\
      TLinFormer   512-512-0.382 & 129.1  & 51.9  & 34.4  & 28.7  & 26.2 & 24.6 & 23.6 & 22.8 & 22.4 & 21.9 \\
      TLinFormer   512-512-0.5   & 127.9  & 52.1  & 34.5  & 28.9  & 26.3 & 24.7 & 23.7 & 22.9 & 22.3 & 21.9 \\
      TLinFormer   512-512-0.618 & 130.1  & 52.4  & 34.4  & 28.8  & 26.2 & 24.8 & 23.6 & 23.0 & 22.4 & 21.9 \\
      \bottomrule
  \end{tabular}
  }
\end{table}

\subsection{Inference Results and Analysis}

\subsubsection{Testing Methodology}
\begin{enumerate}
    \item \textbf{Precondition:} Cache is always enabled.
    \item \textbf{Environment Initialization:} Before each test run, we clear the GPU memory by calling \texttt{torch.cuda.empty\_cache()} to ensure that each test starts from a clean and consistent initial state, eliminating interference from caching.

    \item \textbf{Incremental Sequence Length:} We start with a small initial sequence length (e.g., $N=1$) and then progressively increase the sequence length $N$ by a fixed step (e.g., 10,000 tokens).
    
    \item \textbf{Inference and Timing:} For each initial sequence length $N$, we generate a random integer tensor of shape $(1, N)$ as input. This tensor is fed into the model to generate 6 new tokens. We measure and record the time and cache consumption required for the model to generate each token. We always select the first token (cache miss, equivalent to cache off) and the third token (cache hit) for detailed analysis.

    \item \textbf{Determining Maximum Sequence Length:} After clearing the model's cache, we continue to increase $N$ and repeat step 3 until the model fails to complete inference due to an Out of Memory (OOM) error on the GPU.
\end{enumerate}

\subsubsection{Analysis of Inference Time Complexity (Figures a, b)}
As shown in Figure~\ref{fig:inference_analysis}(a), the inference latency of the standard Transformer baseline exhibits an approximately quadratic ($\mathcal{O}(N^2)$) trend as the sequence length $N$ increases. This phenomenon reveals a gap between theory and practice: although the KV cache reduces the theoretical computational complexity to $\mathcal{O}(N)$, in real-world applications, the performance bottleneck shifts from floating-point operations (FLOPs) to \textbf{memory access (Memory IO)}.

In contrast, TLinFormer (Figure~\ref{fig:inference_analysis}(b)) demonstrates a unique \textbf{dual-mode performance characteristic} (as discussed in Section~\ref{sec:complexity}). The ``peak'' pattern observed in the figure is a direct manifestation of its caching mechanism. Each data point in this test measures the time taken to generate the first token for a \textbf{new, longer initial sequence}, which invariably triggers a \textbf{Cache Miss}. Therefore, the \textbf{peaks} in the plot accurately depict the cache miss costs at various sequence lengths, constituting the \textbf{upper bound} of performance overhead.

This pattern clearly demonstrates the superior scalability of TLinFormer: its performance upper bound (on a cache miss) grows with a \textbf{low-slope linear trend} with respect to the sequence length, far outperforming the baseline's quadratic trend. More critically, the \textbf{valleys} between the peaks represent the cost of generating subsequent tokens during a cache hit state, forming the \textbf{lower bound} of the performance overhead. As illustrated, this lower bound also exhibits a \textbf{linear growth with an extremely low slope}. This is in perfect agreement with our theoretical analysis, which posits that the cost on a \texttt{Cache Hit} is dominated by the interaction of a few layers with the long history, resulting in a complexity of $\mathcal{O}(N)$ but with a very small constant factor. It is this highly efficient linear behavior under both modes that fundamentally explains TLinFormer's overwhelming advantage in long-sequence inference.

\subsubsection{KV Cache Mechanism Efficiency Comparison (Figures c, d)}

To quantify the actual benefits of caching, we compared the inference speedup ratio of both models during cache hits versus cache misses. The results are highly informative:

\noindent\textbf{Baseline Model (Figure c):} As shown in Figure~\ref{fig:inference_analysis}(c), the speedup ratio for the standard model peaks at only 1.26x and rapidly decays to nearly 1.0 as the sequence grows, meaning that in long-sequence scenarios, the KV cache becomes almost \textbf{completely ineffective}. The root cause is the reliance on \texttt{torch.cat} in a naive implementation. This operation needs to reallocate memory and copy the entire, ever-growing KV cache at each generation step, leading to a severe memory bandwidth bottleneck whose overhead almost completely negates the savings from re-computation.

\noindent\textit{Note on Pre-allocation Strategy:} Although this issue can be mitigated with engineering tricks like pre-allocating a larger memory space (pre-allocation), this essentially comes at the cost of higher static memory usage. To ensure a fair comparison with TLinFormer at the algorithmic level, the baseline model in this paper does not employ such additional engineering optimizations.

\noindent\textbf{TLinFormer (Figure d):} As shown in Figure~\ref{fig:inference_analysis}(d), the inference speedup ratio of TLinFormer shows a strong positive correlation with the initial sequence length (N).
As the sequence length grows from zero to about 600,000, the speedup ratio steadily climbs from about 2x to a plateau of about 6x.
When the sequence length exceeds 600,000, the cache efficiency is further enhanced, eventually reaching a peak speedup of over \textbf{10x} on long sequences (N > 800,000).
This indicates that the longer the context, the more significant the performance gain from TLinFormer's caching mechanism.

\subsubsection{Memory Footprint and Supported Sequence Length (Figure e)}

Figure~\ref{fig:inference_analysis}(e) compares the cache memory usage of the two models. Although both exhibit linear growth, consistent with theoretical expectations, the key difference lies in the \textbf{slope of the growth}. The memory usage of the baseline model (`Base 1K`) increases very rapidly. In contrast, TLinFormer's curve is much flatter, demonstrating its extremely high memory efficiency. On the same hardware, TLinFormer supports a maximum sequence length of \textbf{over 1 million tokens}, far exceeding the baseline model's \textbf{120,000 tokens}.

\subsubsection{Overall Inference Speedup Ratio (Figure f)}

\begin{figure}[H]
    \centering

    \begin{subfigure}[b]{0.46\textwidth}
        \centering
        \includegraphics[width=\linewidth]{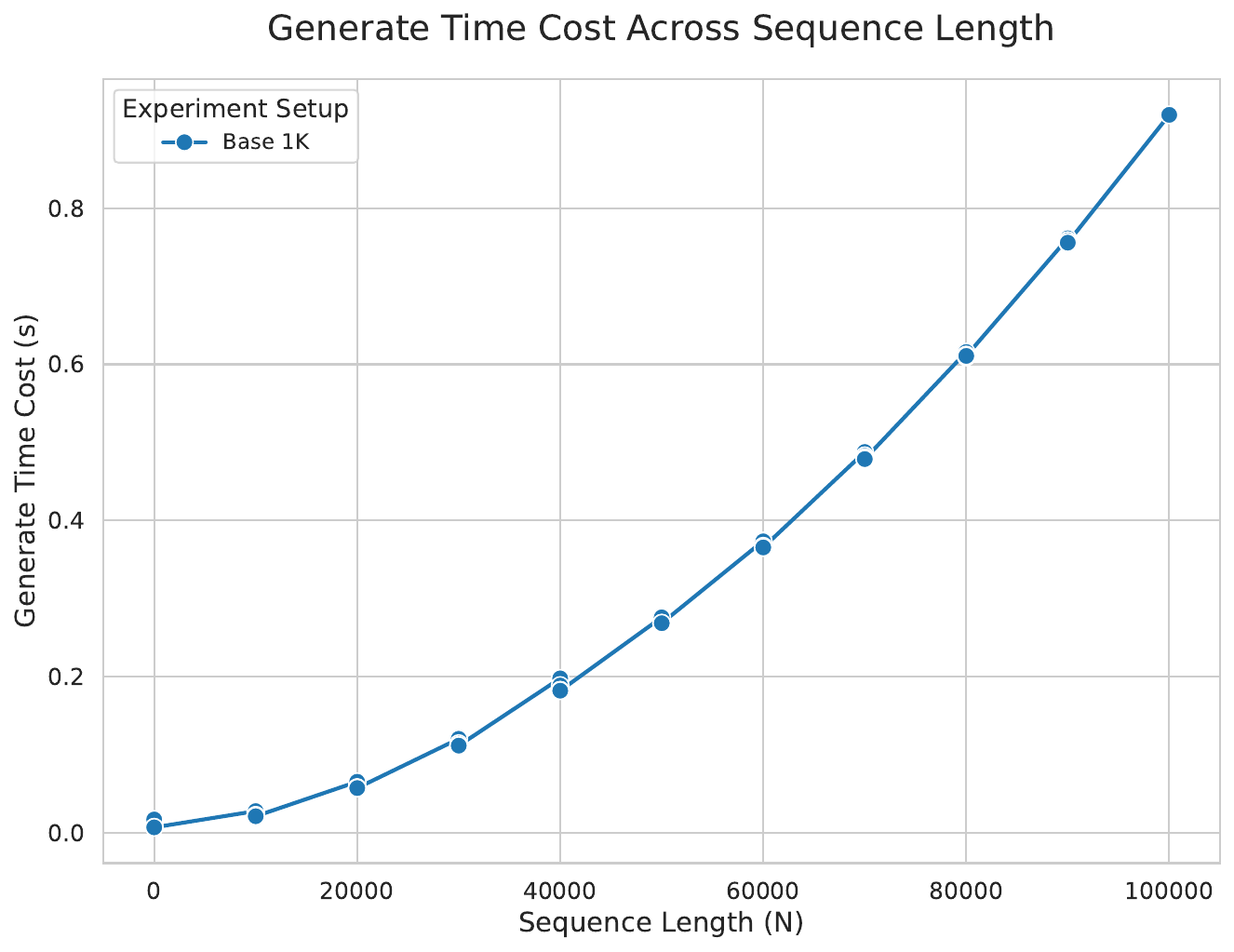}
        \caption{Baseline model inference latency}
        \label{subfig:baseline_latency}
    \end{subfigure}%
    \hfill
    \begin{subfigure}[b]{0.46\textwidth}
        \centering
        \includegraphics[width=\linewidth]{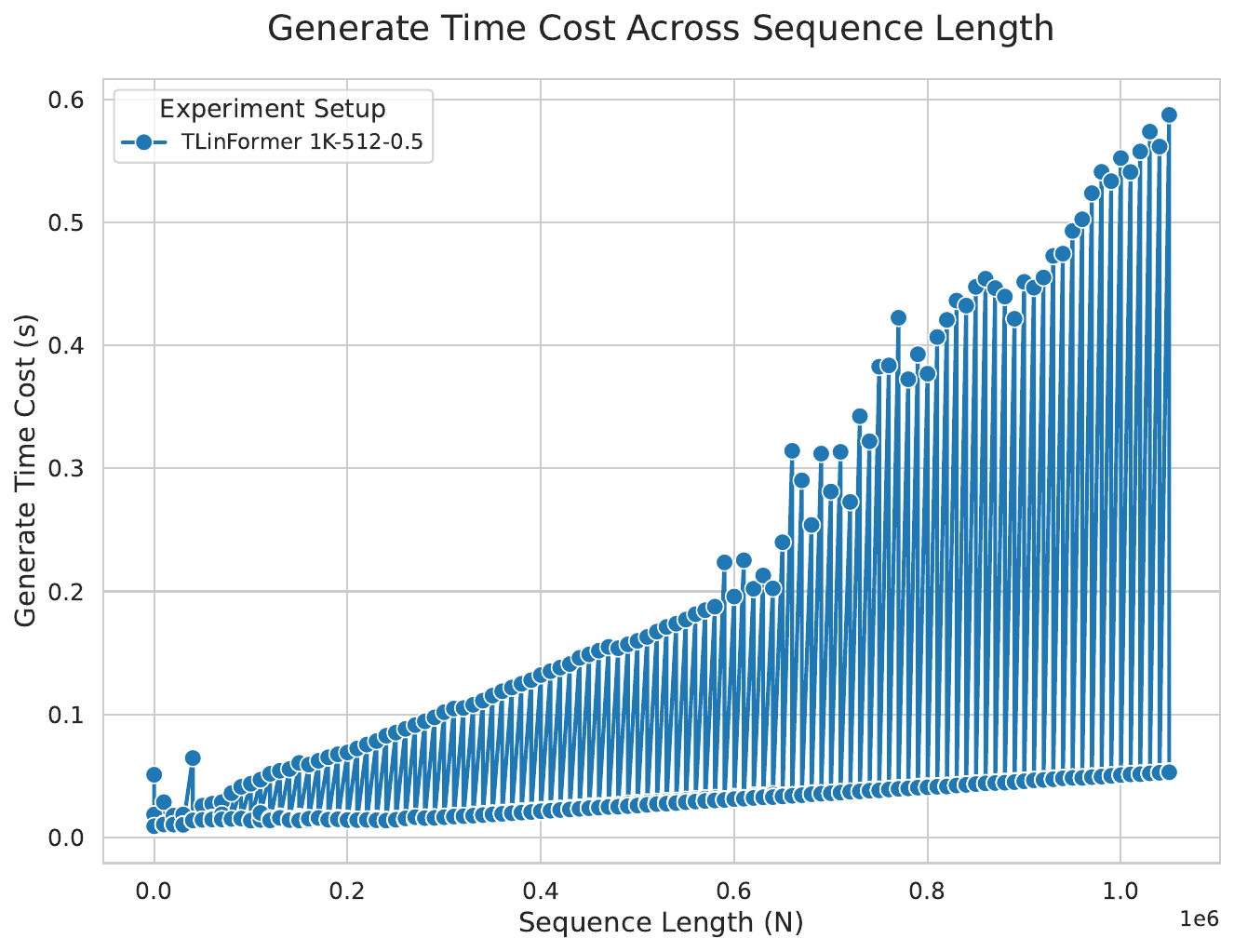}
        \caption{TLinFormer inference latency}
        \label{subfig:tlinformer_latency}
    \end{subfigure}

    \vspace{1.5em}

    \begin{subfigure}[b]{0.46\textwidth}
        \centering
        \includegraphics[width=\linewidth]{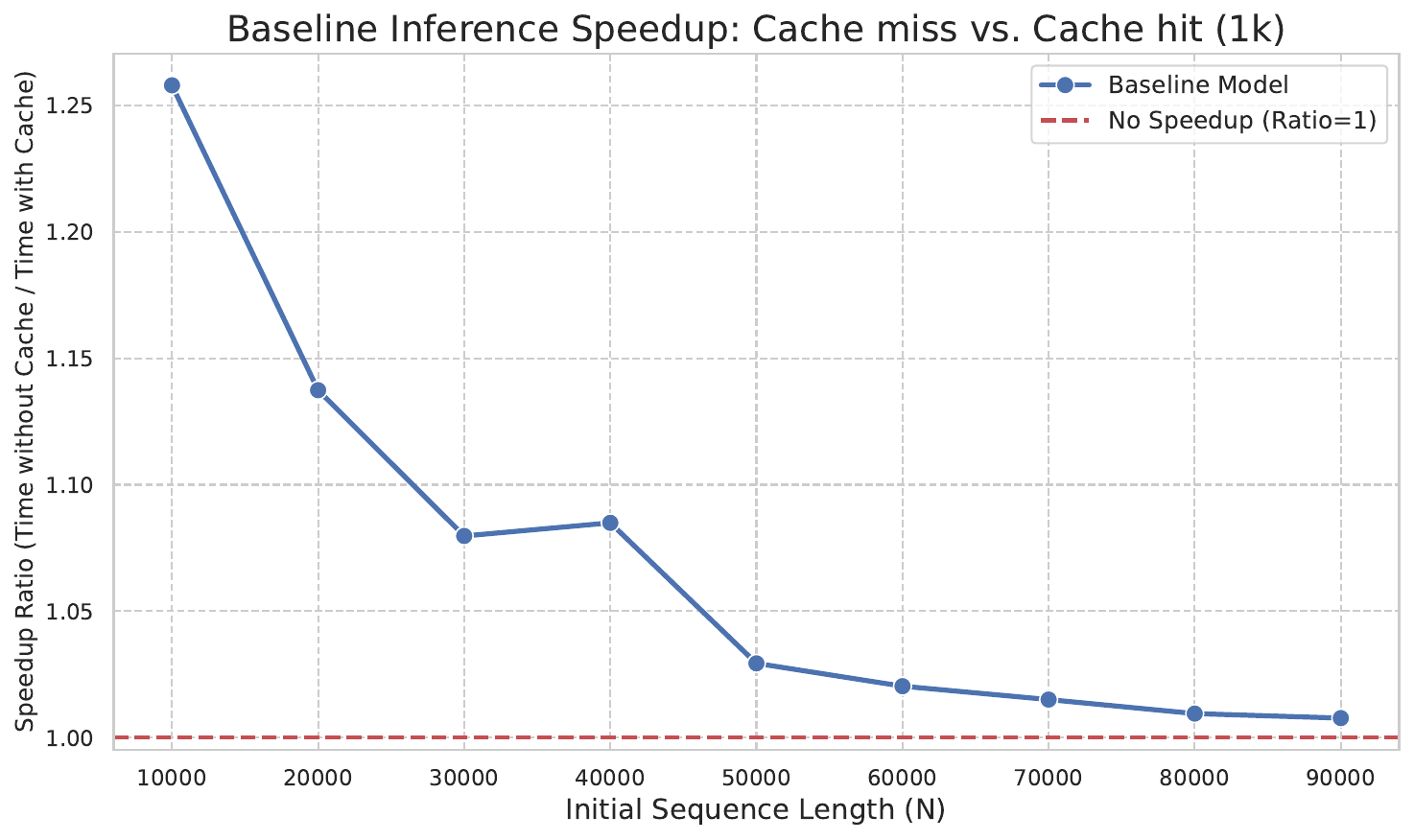}
        \caption{Baseline model inference speedup ratio}
        \label{subfig:baseline_speedup}
    \end{subfigure}%
    \hfill
    \begin{subfigure}[b]{0.46\textwidth}
        \centering
        \includegraphics[width=\linewidth]{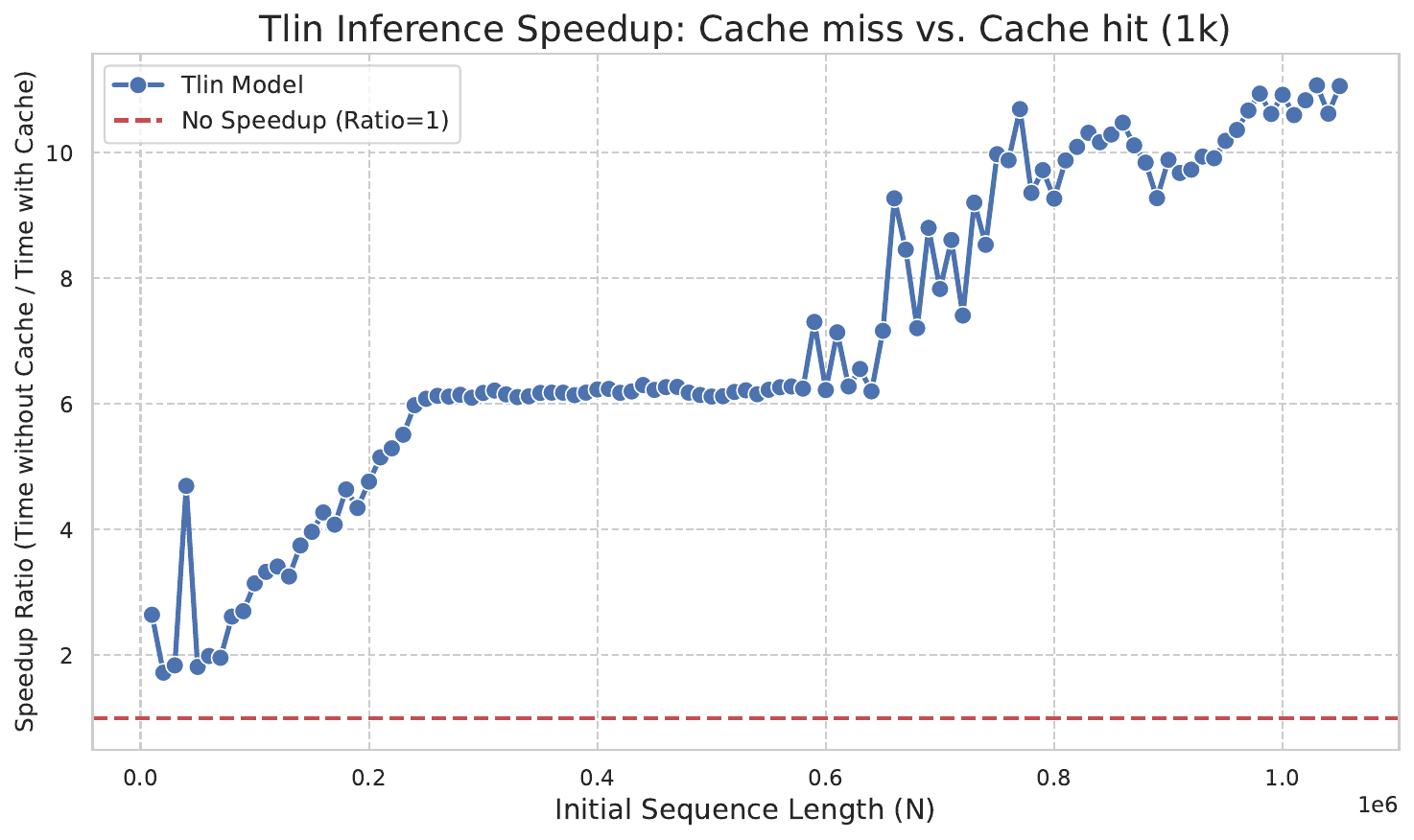}
        \caption{TLinFormer inference speedup ratio}
        \label{subfig:tlin_speedup}
    \end{subfigure}

    \vspace{1.5em}

    \begin{subfigure}[b]{0.46\textwidth}
        \centering
        \includegraphics[width=\linewidth]{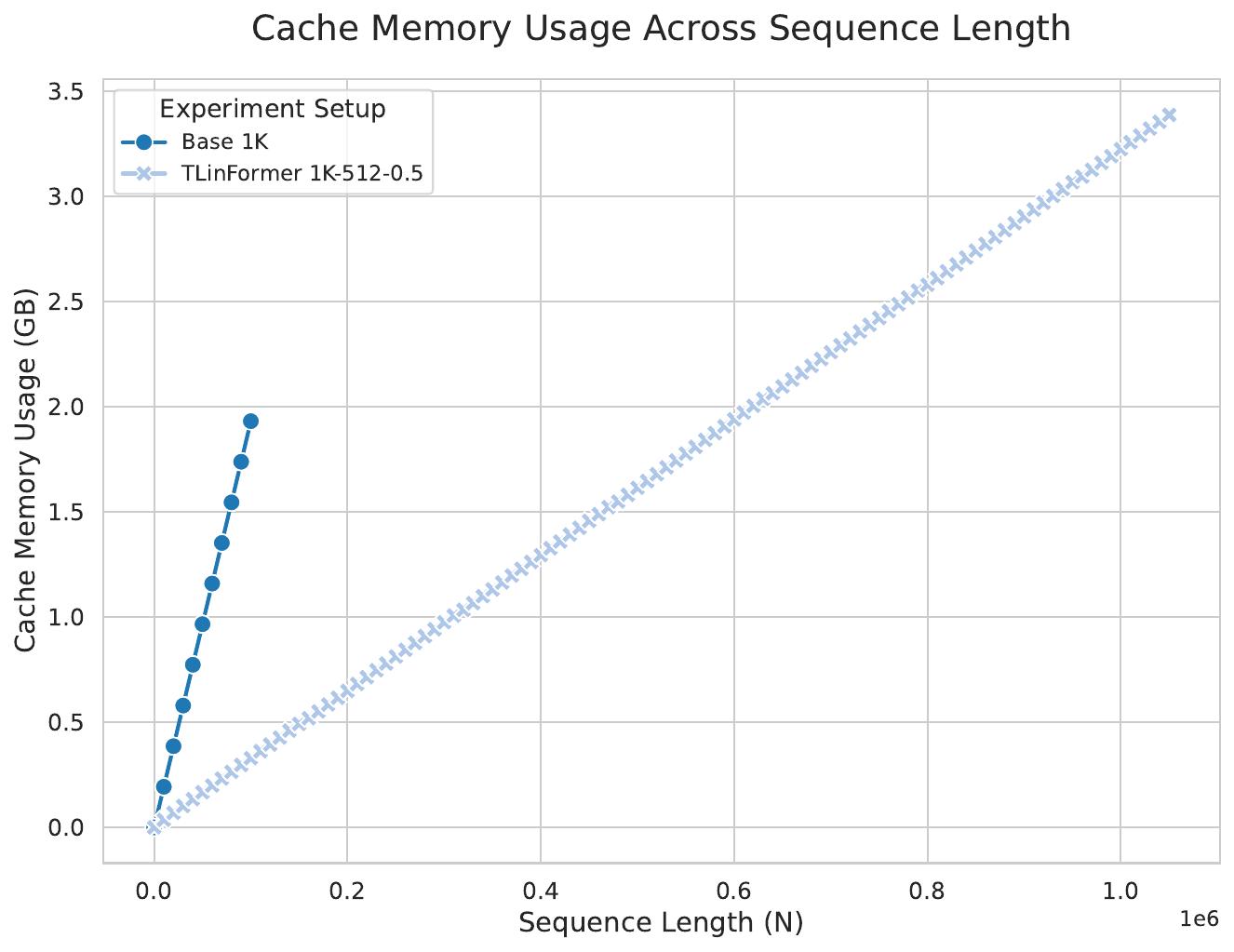}
        \caption{Cache memory usage of TLinFormer vs. baseline}
        \label{subfig:cache_memory}
    \end{subfigure}%
    \hfill
    \begin{subfigure}[b]{0.46\textwidth}
        \centering
        \includegraphics[width=\linewidth]{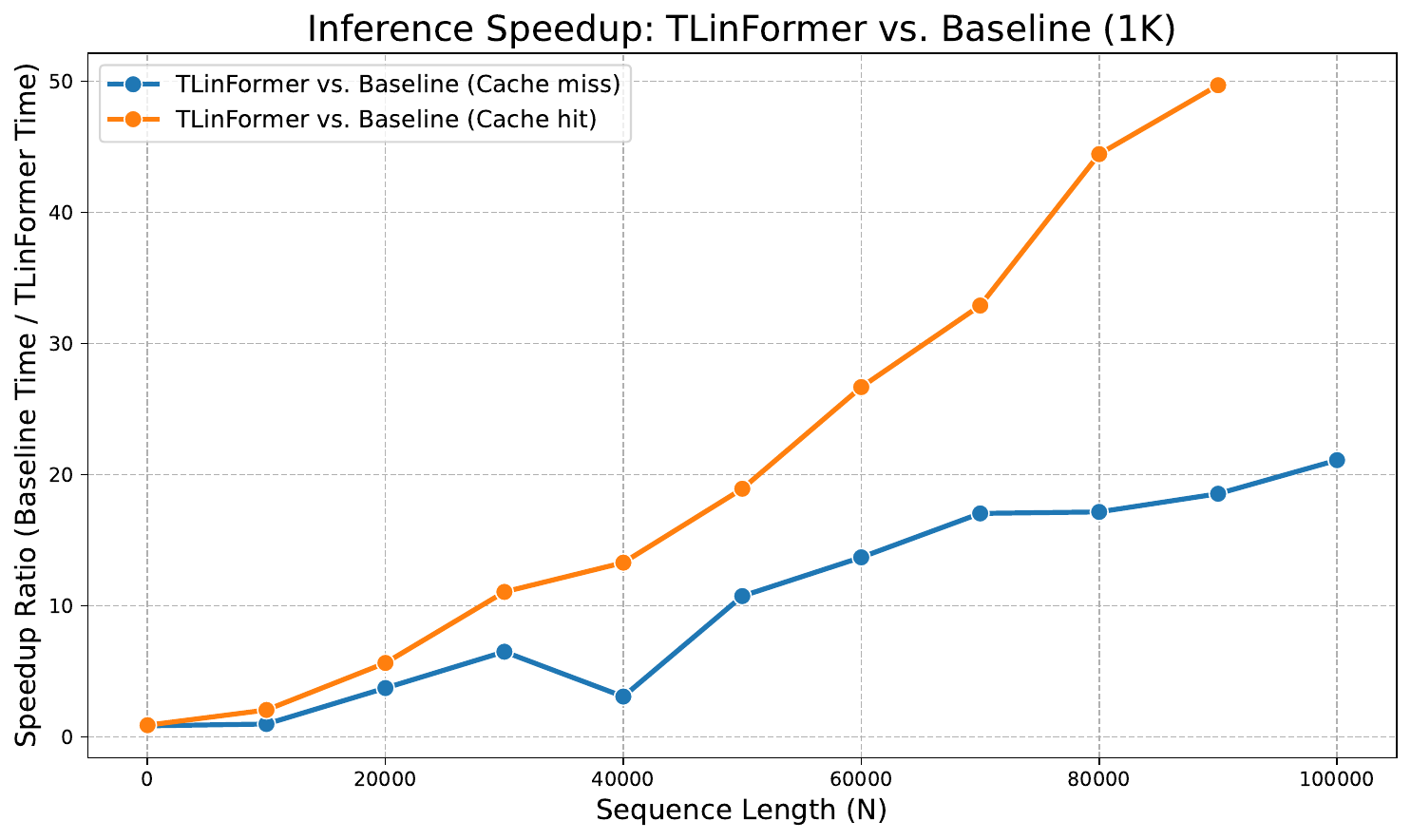}
        \caption{Ratio of TLinFormer to baseline inference time}
        \label{subfig:time_ratio}
    \end{subfigure}

    \caption{Comparison of inference performance and cache efficiency. (a) Baseline model latency shows super-linear growth. (b) TLinFormer's latency pattern demonstrates excellent scalability. (c) Due to a memory bandwidth bottleneck (caused by \texttt{torch.cat} operations), the utility of the baseline model's cache diminishes as sequence length increases. (d) TLinFormer's architecture effectively avoids this bottleneck, showing significant speedup. (e) TLinFormer's cache memory footprint is superior to the baseline's. (f) TLinFormer's inference time is significantly better than the baseline's.}
    \label{fig:inference_analysis}
\end{figure}

Figure~\ref{fig:inference_analysis}(f) visually demonstrates TLinFormer's \textbf{order-of-magnitude speedup advantage} in end-to-end inference compared to the baseline model.
Near the maximum sequence length that the baseline model can handle, TLinFormer achieves up to a \textbf{53x} (cache hit) and \textbf{20x} (cache miss) performance leap in inference speed.

More importantly, the speedup ratio curve shows an approximately \textbf{linear growth trend}, which is not coincidental. It fundamentally stems from the difference in complexity between the two architectures: the gap between TLinFormer's $\mathcal{O}(N)$ complexity and the baseline's practical $\mathcal{O}(N^2)$ complexity systematically widens as the sequence length $N$ increases. Therefore, it can be definitively inferred that for even longer sequences beyond the baseline's processing capability, TLinFormer's speedup effect will continue to grow.

\subsection{Conclusion}

In summary, TLinFormer, through its innovative architectural design, not only achieves linear complexity in computation and memory at a theoretical level but also successfully optimizes the memory bandwidth bottleneck of traditional Transformers in practice. The experimental data fully demonstrates that under the same hardware conditions, TLinFormer can support \textbf{longer sequence lengths} with \textbf{lower inference latency} and \textbf{less memory consumption}, showcasing its immense potential as a next-generation model for long sequences.

\section{Compression as a Path to Intelligence}
\label{sec:compression_as_intelligence}

In this chapter, we engage in an exploratory discussion, re-examining the TLinFormer architecture from the perspective of information compression. Although standard Transformers and their variants have achieved remarkable success across various tasks, the path to Artificial General Intelligence (AGI) still seems unclear. We argue that a potential limiting factor lies in their method of information processing: standard Transformers handle information through an observation window that expands with sequence length, which is more akin to an "exhaustive" association rather than an efficient abstraction and compression of information. Meanwhile, many sparse variants designed for efficiency achieve this by sacrificing Full Context Awareness, which fundamentally undermines the precondition for the model to perform deep compression and abstraction on complete information.

In contrast to the methods above, the design philosophy of TLinFormer embodies the idea of "forced compression." By confining its core computations to a fixed-length observation window while ensuring this window receives input from the entire historical context, we effectively compel the model, in order to complete a given language task, to learn how to distill and compress an infinitely growing history of information into a fixed-size, high-information-density state representation. The model can no longer "cut corners" by expanding its observation scope to find answers; instead, it must engage in deeper semantic understanding and abstraction to generate accurate predictions.

Although TLinFormer is just a small step in this direction, its design philosophy of forced compression offers a glimmer of hope on the path toward Artificial General Intelligence (AGI).

\section{Conclusion and Future Work}
\label{sec:conclusion}

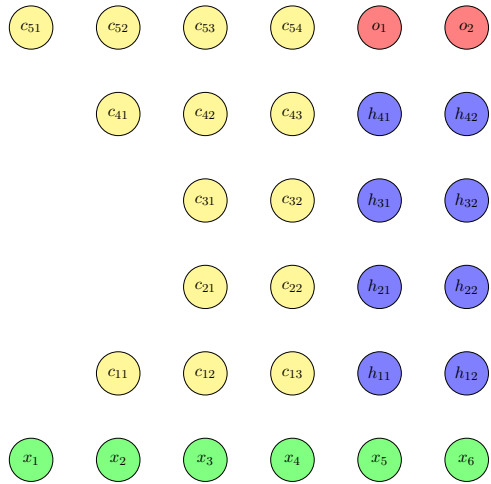
\begin{figure}[H]
    \centering
    \resizebox{0.4\linewidth}{!}{\begin{tikzpicture}[
    neuron/.style={circle, draw, minimum size=1cm},
    input/.style={neuron, fill=green!50},
    hidden/.style={neuron, fill=blue!50},
    output/.style={neuron, fill=red!50},
    memory/.style={neuron, fill=orange!50}, 
    context/.style={neuron, fill=yellow!50}, 
    arrow/.style={->, >=latex, thick}, 
]
    
    \def \numX {6}
    \def \numXc {3}
    \def \numXg {2}
    \def \numCH {2}
    \def \numXH {2}
    \def \numO {2}

    \genLeftAlignedNodeLayer{X}{0}{0}{\numX}{input}{x}{}

    \genLeftAlignedNodeLayer{C1}{2}{2}{3}{context}{c}{1}
    \genLeftAlignedNodeLayer{C2}{4}{4}{2}{context}{c}{2}
    \genLeftAlignedNodeLayer{C3}{4}{6}{2}{context}{c}{3}
    \genLeftAlignedNodeLayer{C4}{2}{8}{3}{context}{c}{4}
    \genLeftAlignedNodeLayer{C5}{0}{10}{4}{context}{c}{5}

    \genLeftAlignedNodeLayer{H1}{8}{2}{\numXH}{hidden}{h}{1}
    \genLeftAlignedNodeLayer{H2}{8}{4}{\numXH}{hidden}{h}{2}
    \genLeftAlignedNodeLayer{H3}{8}{6}{\numXH}{hidden}{h}{3}
    \genLeftAlignedNodeLayer{H4}{8}{8}{\numXH}{hidden}{h}{4}

    \genLeftAlignedNodeLayer{O}{8}{10}{\numO}{output}{o}{}
    





\end{tikzpicture}}
    \caption{A potential AE-inspired connection architecture.}
    \label{fig:a_vae_like_causal_transformer}
\end{figure}

In this paper, we have abandoned the mainstream paradigm of attention approximation and returned to the first principles of connectionism. Starting from the topological structure of information flow, we proposed a novel Exact, Full Context-Aware linear attention architecture—TLinFormer. By reconstructing neuron connections from the bottom up, TLinFormer builds an ideal information flow pattern that achieves strict linear computational complexity while ensuring both computational exactness and full context reachability. Experiments have shown that this approach offers a new path to resolving the inherent performance-efficiency trade-off in existing efficient Transformers, providing a robust and scalable solution for long-sequence modeling and significantly lowering the hardware barrier for its applications.

Despite the great potential demonstrated by TLinFormer, this study has certain limitations. Our validation was primarily conducted on a small-scale (~41M parameters) model. Therefore, a core future work is to scale TLinFormer up to billions or even larger parameter counts and to test its generalization capabilities on a more diverse range of downstream tasks.

Furthermore, the connectionist perspective proposed in this paper opens up several exciting directions for future architectural innovation. First, as a foundational attention framework, TLinFormer's innovations are orthogonal to and compatible with optimization techniques like Mixture-of-Experts (MoE). Combining TLinFormer's linear attention with the parameter efficiency of MoE is a highly promising direction for building next-generation large language models. Second, more complex connection topologies can be explored. For instance, Figure~\ref{fig:a_vae_like_causal_transformer} shows an architecture inspired by autoencoders (AEs), which learns more abstract context representations by introducing an information bottleneck, potentially enhancing model capabilities. We can also investigate strategies for dynamically adjusting the $\Woh$ and $\Wog$ window sizes during inference based on the context to achieve adaptive optimization of computational load (as discussed in~\ref{sec:discussion_of_windows}).

Finally, the core idea of this paper leads to a more profound consideration: MLPs grant models their basic representation power through full connectivity in the feature dimension (D); Transformers extend this to the sequence dimension (L), unlocking powerful context modeling capabilities. A natural and exciting inference is whether we can design a higher-order, generalized fully-connected mechanism that operates on multi-dimensional data tensors (e.g., of shape \texttt{[B, M, L, D]})? We speculate that developing a computationally efficient and feasible High-dimensional Tensorial Attention could be a key step toward more general and powerful artificial intelligence models. We believe that the connectivity-centric design philosophy proposed in this paper opens a promising new path for future research in efficient and powerful sequence modeling.

\section*{Code Availability}
The source code for this paper is available at \url{https://github.com/simonFelix-Ai/TLinFormer}. The code is dual-licensed for academic and commercial use.

\bibliographystyle{plain}
\bibliography{references} 

\appendix 
\section{Detailed Derivation of Computational Complexity}
\label{app:complexity_derivation}

This appendix derives the full expressions for the model's computational complexity under the two cases of a \textbf{Cache Miss} and a \textbf{Cache Hit}, corresponding to equations~\eqref{eq:total_cost_full_cache_miss} and~\eqref{eq:total_cost_full_cache_hit} in the main text.

\subsection{Cache Miss}
\begin{enumerate}
    \item \textbf{Computational Cost of the Left Window (Historical Context)}:
    \begin{itemize}
        \item \textbf{First-layer cross-attention}: The query sequence from the context window attends to the entire history. Cost: $D \cdot (N - \Wog) \cdot \Woh$.
        \item \textbf{Intermediate self-attention layers ($H$ layers)}: Self-attention is performed within the context window of size $\Woh$. Cost: $H \cdot D \cdot \Woh^2$.
        \item \textbf{Final-layer cross-attention (dimensionality restoration)}: The entire history attends to the processed context window to restore the original sequence length. Cost: $D \cdot (N - \Wog) \cdot \Woh$.
        \item \textit{\textbf{Total Cost of the Left Window ($C_{\text{left}}$):}}
        \[
        C_{\text{left}} = 2 D (N - \Wog) \Woh + H D \Woh^2
        \]
    \end{itemize}
\item \textbf{Computational Cost of the Right Window (Generation Area)}:
\begin{itemize}
    \item \textbf{First cross-attention with the history}: The generation window attends to the entire history. Cost: $D \cdot \Wog \cdot (N - \Wog)$.
    \item \textbf{Intermediate cross-attention with the context ($H+1$ layers, including the final output layer)}: The generation window attends to the processed context window. Cost: $(H + 1) \cdot D \cdot \Wog \cdot \Woh$.
    \item \textbf{Causal self-attention (across all $H+2$ layers, including the final output layer)}: Causal self-attention is performed within the generation window. Cost: $(H + 2) \cdot D \cdot \Wog^2$.
    \item \textit{\textbf{Total Cost of the Right Window ($C_{\text{right}}$):}}
    \[
    C_{\text{right}} = D \Wog (N - \Wog) + (H + 1) D \Wog \Woh + (H+2)D \Wog^2
    \]
\end{itemize}

\item \textbf{Derivation of Total Computational Cost ($T$):}
The total cost is the sum of the costs for the two windows, $T = C_{\text{left}} + C_{\text{right}}$.
\begin{align*}
    T &= \left[ 2 D (N - \Wog) \Woh + H D \Woh^2 \right] \\
      &\quad + \left[ D \Wog (N - \Wog) + (H + 1) D \Wog \Woh + (H+2)D \Wog^2 \right] \\
    \intertext{Step 1: Expand all terms}
    &= \left( 2DN\Woh - 2D\Wog\Woh + HD\Woh^2 \right) \\
      &\quad + \left( DN\Wog - D\Wog^2 + HD\Wog\Woh + D\Wog\Woh + HD\Wog^2 + 2D\Wog^2 \right) \\
    \intertext{Step 2: Combine like terms}
    &= 2DN\Woh + DN\Wog - D\Wog\Woh + HD\Wog\Woh + HD\Woh^2 + HD\Wog^2 + D\Wog^2 \\
    \intertext{Step 3: Factor out the common term $D$ and regroup by variables $N$ and $H$}
    &= D \left[ N(2\Woh + \Wog) - \Wog\Woh + H\Wog\Woh + H\Woh^2 + H\Wog^2 + \Wog^2 \right] \\
    \intertext{Step 4: Final organized form (grouping terms related to $H$)}
    &= D \left[ N(2\Woh + \Wog) + H(\Woh^2 + \Wog^2 + \Wog\Woh) + \Wog^2 - \Wog\Woh \right]
\end{align*}
    Derivation complete.
\end{enumerate}

\subsection{Cache Hit}

\begin{enumerate}
    \item \textbf{Computational Cost of the Left Window (Historical Context)}:
    \begin{itemize}
        \[
        C_{\text{left}} = 0
        \]
    \end{itemize}
    \item \textbf{Computational Cost of the Right Window (Generation Area)}:
    \begin{itemize}
        \item \textbf{First cross-attention with the history}: The generation window attends to the original, entire history, but only the last token participates in the computation. Cost: $D \cdot (N - \Wog)$.
        \item \textbf{Intermediate cross-attention with the context ($H+1$ layers, including the final output layer)}: The generation window attends to the processed context window. Again, only the last token participates in the computation. Cost: $(H + 1) \cdot D \cdot \Woh$.
        \item \textbf{Causal self-attention (across all $H+2$ layers, including the final output layer)}: Causal self-attention is performed within the generation window. Cost: $(H + 2) \cdot D \cdot \Wog^2$.
        \item \textit{\textbf{Total Cost of the Right Window ($C_{\text{right}}$):}}
        \[
        C_{\text{right}} = D (N - \Wog) + (H + 1) D \Woh + (H+2)D \Wog^2
        \]
    \end{itemize}

\item \textbf{Derivation of Total Computational Cost ($T$):}
The total cost is the sum of the costs for the two windows, $T = C_{\text{left}} + C_{\text{right}}$.
\begin{align*}
    T &= D (N - \Wog) + (H + 1) D \Woh + (H+2)D \Wog^2
\end{align*}
    Derivation complete.
\end{enumerate}

\end{document}